\documentclass[conference]{IEEEtran}
\usepackage{cite}
\usepackage{amsmath,amssymb,amsfonts}
\usepackage{algorithmic}
\usepackage{graphicx}
\usepackage{textcomp}
\usepackage{xcolor}
\usepackage[hyphens]{url}
\usepackage{fancyhdr}
\usepackage[bookmarks=true,breaklinks=true,letterpaper=true,colorlinks,citecolor=blue,linkcolor=blue,urlcolor=blue]{hyperref}

\usepackage{subfig}
\usepackage{makecell}
\usepackage{algorithm}
\usepackage{algorithmic}
\usepackage{ulem}
\usepackage{pifont}
\usepackage{multirow}

\title{An Asynchronous Multi-core Accelerator for SNN inference} 

\DeclareRobustCommand*{\IEEEauthorrefmark}[1]{%
  \raisebox{0pt}[0pt][0pt]{\textsuperscript{\footnotesize #1}}%
}
\author{
    \IEEEauthorblockN{
        Zhuo Chen\IEEEauthorrefmark{1}, 
        De Ma\IEEEauthorrefmark{1}, 
        Xiaofei Jin\IEEEauthorrefmark{1}\IEEEauthorrefmark{2},
        Qinghui Xing\IEEEauthorrefmark{1},
        Ouwen Jin\IEEEauthorrefmark{1},
        Xin Du\IEEEauthorrefmark{1},
        Shuibing He\IEEEauthorrefmark{1} and
        Gang Pan\IEEEauthorrefmark{1}
    }
    \IEEEauthorblockA{
        \IEEEauthorrefmark{1}Zhejiang University\\
        \IEEEauthorrefmark{2}Zhejiang Lab\\
            \{chenzhuocs, made\}@zju.edu.cn, 
            jinxf@zhejianglab.com,
            \{xingqh, jinouwen\}@zju.edu.cn,\\jsjduxin@gmail.com, \{heshuibing, gpan\}@zju.edu.cn
    }
}

\begin{document}
\maketitle
\pagestyle{plain}


\begin{abstract}

  Spiking Neural Networks (SNNs) are extensively utilized in brain-inspired computing and neuroscience research. To enhance the speed and energy efficiency of SNNs, several many-core accelerators have been developed. However, maintaining the accuracy of SNNs often necessitates frequent explicit synchronization among all cores, which presents a challenge to overall efficiency. In this paper, we introduce an asynchronous architecture that eliminates inter-core synchronization, thereby enabling fast and energy-efficient SNN inference with commendable scalability. The core concept is to leverage the dependency of neuromorphic cores predetermined at compile time. We design a scheduler for each core to monitor the running state of its dependencies and control the core to safely proceed to the next timestep without waiting for other cores to complete their tasks. This approach eliminates the need for global synchronization, allowing our architecture to minimize core waiting time in the face of inherent core and time imbalances in SNN workloads. Comprehensive evaluations using five SNN workloads demonstrate that the proposed architecture achieves a 1.86x speedup and 1.55x energy efficiency compared to the state-of-the-art synchronization architectures.

\end{abstract}

\section{Introduction}

\begin{figure}
    \centering
    \subfloat[ANN neuron model.]{\includegraphics[width=0.4\linewidth]{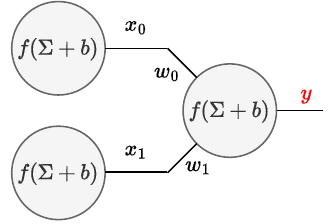}}
    \subfloat[SNN neuron model.]{\includegraphics[width=0.5\linewidth]{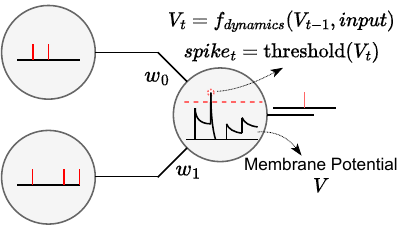}}
    \caption{Different neuron models between ANN and SNN.}
    \label{fig:neuron}
    \vspace{-1.5em}
\end{figure}

Brain-inspired computing, or neuromorphic computing, aims to simulate brain behavior to energy-efficiently achieve artificial intelligence~\cite{roy2019towards}. Spiking neural networks (SNNs)~\cite{maass1997networks} are widely used models for neuromorphic computing and neuroscience research. Rather than conventional artificial neural networks (ANNs) use only rate coding, SNNs are more similar to biological neural networks, using various coding with richer spatial-temporal information. Neurons in SNNs are supposed to model simplified dynamics like biological neurons, which is mathematically equivalent to discrete-updated differential equations (Fig.~\ref{fig:neuron}).

To exploit the great potential in low power consumption of SNNs, several neuromorphic accelerators have been proposed. They can be divided into two categories. The first kind of SNN accelerators~\cite{narayanan2020spinalflow,liu2022sato,lee2022parallel,li2023firefly,yin2022sata} adopts an architecture similar to existing ANN accelerators~\cite{jouppi2017datacenter}, which composes a 2-D systolic array of processing elements (PEs) and a large shared global buffer. The ANN-accelerator-like architecture separates computation units and memory units, which is not suitable for SNNs. On the other hand, a near-memory many-core architecture can dramatically eliminate redundant data movement, which is adopted by some industrial neuromorphic chips, such as TrueNorth~\cite{akopyan2015truenorth}, SpiNNaker~\cite{furber2014spinnaker}, Loihi~\cite{davies2018loihi}, Tianjic~\cite{pei2019towards} and Darwin~\cite{ma2017darwin}. In general, they comprise of many neuromorphic cores with built-in memories for computation, and an on-chip network (NoC) for spike communication. There are many researches~\cite{lee2018flexon,baek2019flexlearn,chen2022gaban,lee2022neurosync} improving performance, flexibility, and energy efficiency. In this work, we focus on this architecture.

Although recent multi-core SNN accelerators can reduce memory access to lower power consumption, they suffer from low utilization caused by periodic all-core synchronization. The synchronization procedure is necessary for the SNN computation mode mentioned before, where each core in an accelerator updates neuron states step-by-step. Each neuromorphic core should wait for all other cores and spike transfers at each timestep to ensure result correctness, wasting considerable time. Many solutions have been proposed to improve SNN accelerator performance by eliminating or relaxing synchronization. Unfortunately, they either require new electronic devices~\cite{xia2019memristive,chen2023open}, or compromise the accuracy of the results~\cite{neil2014minitaur,lee2021neuroengine}, or reduce the synchronization frequency by speculative execution mechanism meanwhile bring rollback overheads~\cite{lee2022neurosync}.

\begin{figure}
    \centering
    \includegraphics[width=0.8\linewidth]{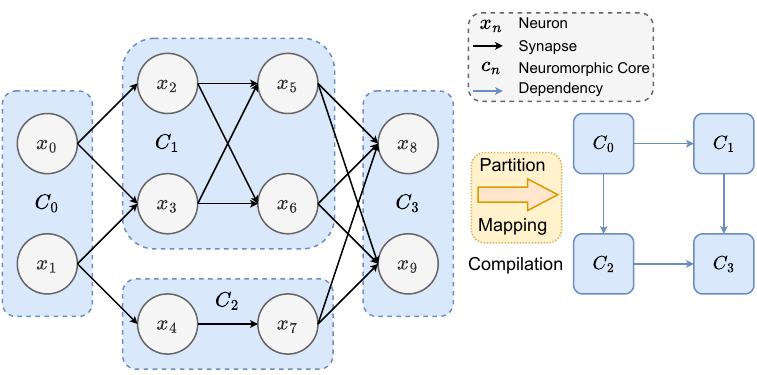}
    \caption{Deploy pipeline on SNN accelerators.}
    \label{fig:compile}
    \vspace{-1.5em}
\end{figure}

In this work, we propose DepAsync to tackle this problem without global synchronization and additional devices, meanwhile keeping exact spike results at each timestep which we called time-accuracy.
We first explain synchronization from the perspective of data dependency among neuromorphic cores in SNN accelerators. Then we introduce core dependency determined at compilation time (Fig.~\ref{fig:compile}) as prior information into time-driven architectures and design a mechanism to dynamically trace running states of dependencies for each core. With static core dependency and dynamic running states, DepAsync can control cores to safely forward to the next timestep, reducing redundant waiting time.
Our asynchronous mechanism eliminates all-core synchronization without speculative execution and is more flexible to make fine-grained control in core-imbalance and time-imbalance workloads, achieving higher speedup and energy efficiency. Our experiments use five different SNNs to evaluate DepAsync. The results show that DepAsync improves performance and energy cost by 1.86x and 1.55x over plain time-driven accelerators, and 1.46x, 1.29x over speculative-execution architectures.

The main contributions of this work are three-fold:
\begin{itemize}
    \item We identify that the synchronization in time-driven SNN accelerators derives from data dependencies between neuromorphic cores and review time-driven architectures by analyzing core dependencies.
    \item We introduce core dependencies as prior information to time-driven SNN hardware accelerators, and propose DepAsync to eliminate all-core synchronization as a solution to the low utilization problem.
    \item We evaluate our architecture with five different SNN workloads and demonstrate its advantages over existing accelerators.    
\end{itemize}

\section{Background}
\subsection{Time-driven SNN Inference}
Neurons in SNNs, acting as biological neurons, keep updating internal states (i.e., membrane potential) and output a spike when the voltage reaches the preset threshold as illustrated in Fig.~\ref{fig:neuron}, which gives SNNs the ability to model richer spatial-temporal information than ANNs. For example, the Leaky Integrate-and-Fire (LIF) model is prevalent in nowadays SNNs because of its good balance between biological plausibility and computing complexity. In mathematics, the internal state updating procedure can be represented as a differential equation like Equation~\ref{equ:1}:
\begin{equation}
    \label{equ:1}
    \tau_m \frac{dV}{dt} = -(V - V_{rst}) + \frac{I}{g_L}
\end{equation}
where $V$ is the membrane potential, $I$ is the external input, and $V_{rst},\tau_m,g_L$ are parameters, representing resetting potential, membrane time constant, and leak conductance, respectively.

However, it is hard and inefficient to calculate exact analytical solutions using digital circuits. Thus, in order to simulate an SNN on digital circuits efficiently, some numerical methods are used in solving differential equations in Equation~\ref{equ:1}, where the differential w.r.t. time $dt$ is usually called \textit{timestep}. Fig.~\ref{fig:compile} shows an example of SNNs, where $x_i$ denotes neurons and the arrows represent connections between them (i.e., synapses) and the overall SNN inference can be written as Algorithm~\ref{alg:1}.

\begin{algorithm}[b]
    \renewcommand{\algorithmicrequire}{\textbf{Input:}}
    \renewcommand{\algorithmicensure}{\textbf{Output:}}
    \caption{SNN inference in the time-driven manner}
    \label{alg:1}
    \begin{algorithmic}[1]
        \REQUIRE neurons $\{x_i\}_{i=1}^N$, synapse weight $\{w_{ij}\}_{N\times N}$
        \REQUIRE threshold $v_{th}$, reset potential $v_{rst}$
        \REQUIRE maximum timestep $t_{max}$
        \ENSURE spikes $\{s_{ij}\}_{N\times t_{max}}$
        \WHILE{$t < t_{max}$}
            \FORALL{$x_i \in \{x_i\}_{i=1}^N$}
                \STATE $x_i.states = \text{neuron\_model}(x_i.states, x_i.acc)$
                \STATE $s_{it} = x_i.states > v_{th}$
                \IF{$s_{it}$}
                    \STATE $x_i.states = v_{rst}$
                    \FORALL{$x_j \in x_i.fanout$}
                        \STATE $x_j.acc = x_j.acc + w_{ij}$
                    \ENDFOR
                \ENDIF
            \ENDFOR
            \STATE $t = t + 1$
        \ENDWHILE
    \end{algorithmic}  
\end{algorithm}

\subsection{Many-core Time-driven SNN Accelerators} \label{sec:many-core}
\begin{figure}[t]
    \centering
    \includegraphics[width=0.6\linewidth]{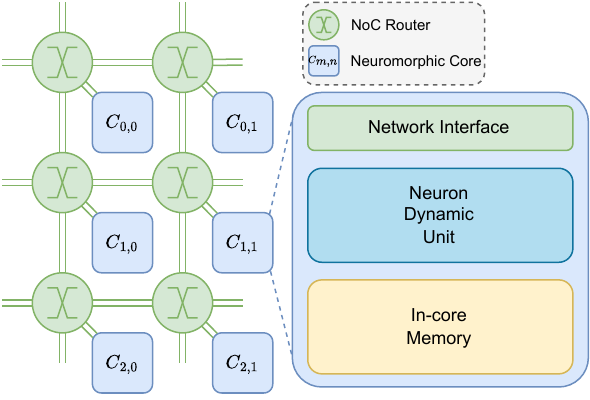}
    \caption{Block diagram of many-core SNN accelerators.}
    \label{fig:snn-accelerator}
    \vspace{-1.5em}
\end{figure}

Compared with GPU-based acceleration~\cite{Wang2022brainpy,hong2022spaic,fang2023spikingjelly} and systolic-array-based SNN accelerators~\cite{narayanan2020spinalflow,liu2022sato,lee2022parallel,yin2022sata}, a custom-designed near-memory architecture can leverage more energy-efficient advantages of SNN as it eliminates intense data movement between global and local memory, and therefore has been adopted by industrial neuromorphic chips~\cite{akopyan2015truenorth,furber2014spinnaker,davies2018loihi,ma2017darwin}. This architecture is usually composed of many individual neuromorphic cores and an on-chip network, as illustrated in Fig.~\ref{fig:snn-accelerator}. The neuron unit in each neuromorphic core calculates neuron state updating step-by-step and the in-core memory is for storing neuron states and synapse weights. For simplicity, we still use the term time-driven accelerators to represent many-core architectures in the subsequent sections without specific clarification.

\begin{figure}
    \centering
    \subfloat[Topology.]{
        \raisebox{.35\height}{
            \includegraphics[width=0.2\linewidth]{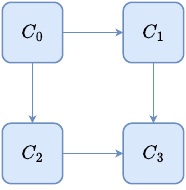}
            \label{fig:topology}
        }
    }
    \subfloat[SNN workflow.]{
        \includegraphics[width=0.45\linewidth]{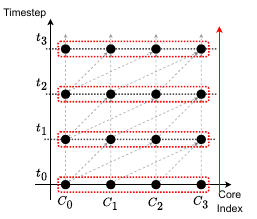}
        \label{fig:time-driven-workflow}
    }
    \caption{SNN inference on time-driven SNN accelerators.}
    \label{fig:time-driven}
    \vspace{-1.5em}
\end{figure}

The primary contribution to accelerating SNN inference mainly comes from exploiting inter-core parallelism within the single timestep. Neuromorphic cores can safely parallelize neuron internal state updating since there is no data dependency at this stage. Then, all cores go through a synchronization stage to wait for spikes generated in this timestep to arrive at destination cores via the NoC. Fig.~\ref{fig:topology} illustrates the topology of a typical SNN inference on a $4\times4$ time-driven accelerator with black arrows representing core dependency, and Fig.~\ref{fig:time-driven-workflow} shows the corresponding SNN inference workflow: cores perform neuromorphic computing (black dots) in parallel before synchronization (in a red rectangle), then work in a timestep-by-timestep manner. The whole system can only simulate SNNs step-by-step, restricted to per-timestep synchronization.

To implement a synchronization mechanism, TrueNorth uses a 1 kHz synchronization trigger signal~\cite{akopyan2015truenorth}. When receiving this tick arrival, the neuromorphic core will be triggered to read the spike buffer and forward to the next timestep. SpiNNaker adopts a similar strategy, defining a timer event by a fixed number of clock cycles~\cite{rhodes2018spynnaker}. It also provides a catch-up mechanism to handle time imbalance in SNN workloads. Loihi~\cite{davies2018loihi} realizes synchronization more adaptively. After neuron state updating, cores in Loihi will send special packages that will drill up spike packages remaining in NoCs. Darwin uses both synchronization methods to give users more options~\cite{ma2017darwin}.


\subsection{Under-utilization Problem}
Although time-driven SNN accelerators dispatch neuron-updating computation to parallel neuromorphic cores, the sparsity nature of spiking data in SNN workloads is not adequately considered. Compared with dense multiply-and-accumulate operations in ANNs, highly sparse spikes make workloads among timesteps and cores become imbalanced, which hinders overall hardware acceleration.

\subsubsection{Imbalance in SNN Workloads}
\begin{figure}
    \centering
    \subfloat[Time-imbalance.]{
        \includegraphics[width=0.5\linewidth]{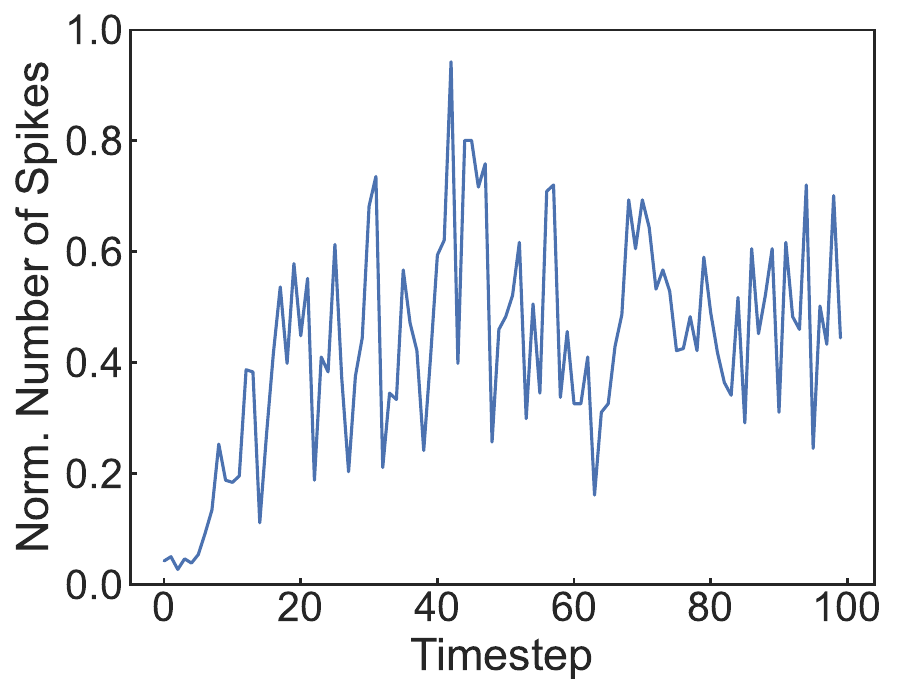}
        \label{fig:time-imbalance}
    }
    \subfloat[Core-imbalance.]{
        \includegraphics[width=0.42\linewidth]{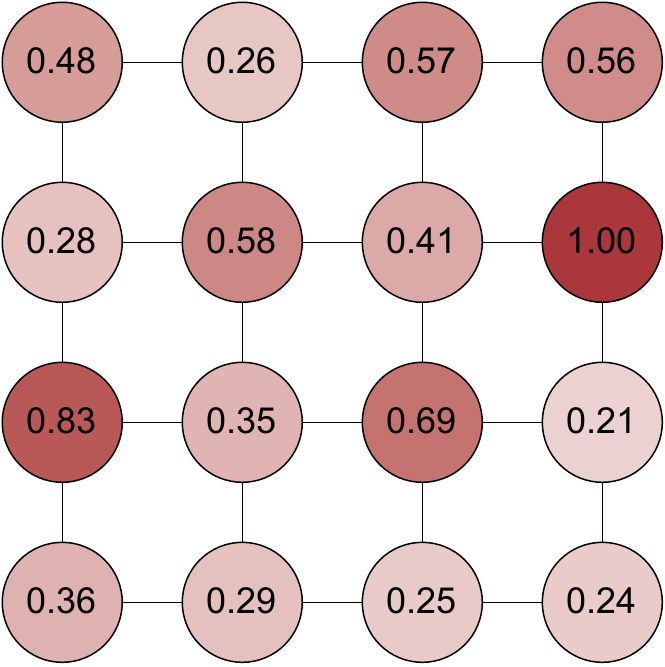}
        \label{fig:core-imbalance}
    }
    \caption{Imbalance in SNN workloads.}
    \label{fig:imbalance}
    \vspace{-1.2em}
\end{figure}

We perform an SNN inference with several workloads to demonstrate workload imbalance. The evaluation details are discussed in Section~\ref{sec:eva}. The SNN runs 100 timesteps on a $4\times4$ accelerator. Fig.~\ref{fig:time-imbalance} shows that generated spikes exhibit significant variability across timesteps; and Fig.~\ref{fig:core-imbalance} renders $4\times4$ neuromorphic cores with different colors and numbers in each circle representing normalized firing rates of each core. Timestep imbalance is from temporal sparsity in SNNs, that is, neurons don't fire at all timesteps. On the other hand, the core imbalance is from spatial sparsity, which means not all neurons fire at each timestep. Thus the total number of spikes is different.

\subsubsection{Under-utilization Caused by Synchronization}
The under-utilization problem in time-driven accelerators has been observed in ~\cite{lee2022parallel,lee2022neurosync,yang2022unicorn}. It's difficult for a rigid synchronization mechanism to simulate imbalanced workloads. Fix synchronization signals cannot be changed dynamically, which is not suitable for temporal-variant spike rates. Synchronization messages after the timestep finish make some cores always wait for the slowest core in the whole system, which cannot tackle inter-core imbalance. What makes it worse is that spike transfer time in NoC grows as the number of cores scales up, and workloads with higher firing rates generate more spikes which cause more NoC congestion. Both of these factors deteriorate the under-utilization problem.

\begin{figure}
    \subfloat[Under-utilization with scale.]{
        \hspace*{-1em}
        \includegraphics[width=0.48\linewidth]{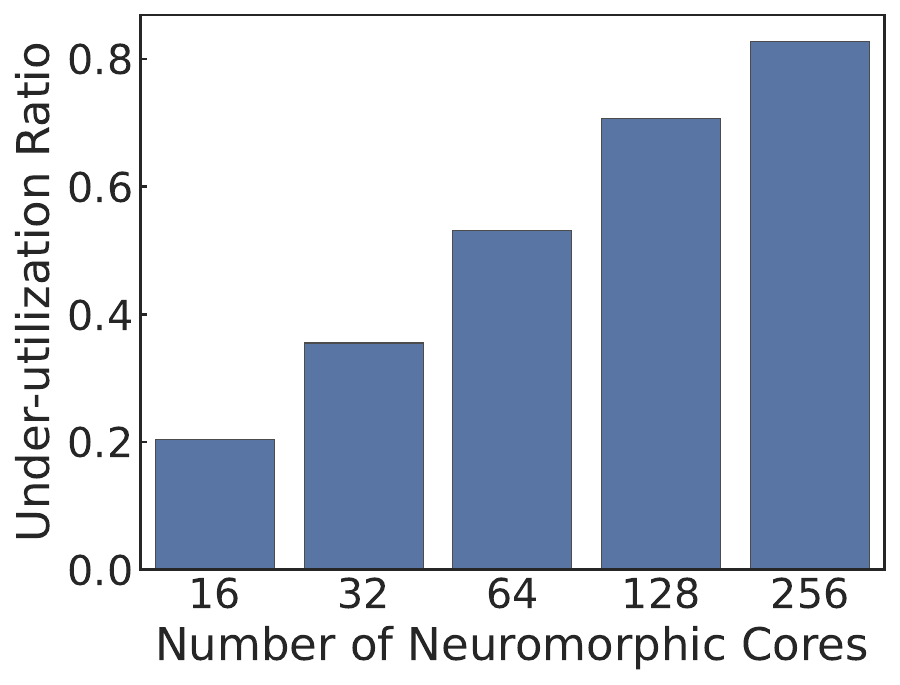}
        \label{fig:core_utilization_scale}
    }
    \subfloat[Under-utilization with firing rate.]{
        \includegraphics[width=0.5\linewidth]{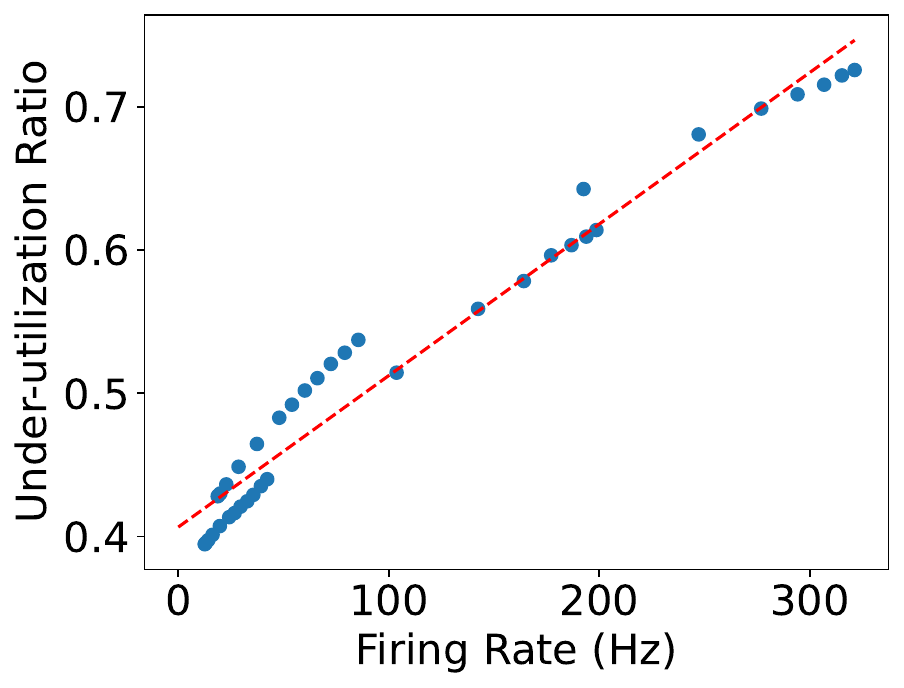}
        \label{fig:core_utilization_fr}
    }
    \caption{Under-utilized cycle ratio of time-driven accelerators.}
    \vspace{-1.2em}
\end{figure}

Fig.~\ref{fig:core_utilization_scale} illustrates that the utilization ratio decreases with the growth of spike transfer time caused by increasing scales. Then, we adjust the average firing rate of SNN workloads. The results in Fig.~\ref{fig:core_utilization_fr} show that the under-utilization problem becomes worse as the firing rate increases, due to more generated spikes slowing down the transmission ability of the NoC.

\subsection{Related Work}
\noindent There are several approaches to reduce synchronization.

Firstly, leverage novel devices. Memristors are efficient emulators of biological neurons and synapses, because of the similarity in using ion migration as the fundamental mechanism~\cite{xia2019memristive}. Analog circuits with memristors can directly simulate neuron dynamics~\cite{stoliar2017leaky}. Furthermore, memristors also leverage physical characteristics to store synapse weights, and even simulate learning rules to support synapse plasticity~\cite{wang2012synaptic}. Thus, timesteps become unnecessary in such total neuromorphic chips. Recently, Chen et al.~\cite{chen2023open} use a passive electrochemical memory (ECRAM) array to improve programming accuracy for analog circuits. However, these emerging devices are still in the research stage, and not as mature as the large-scale integrated circuit technology.

Secondly, get rid of time accuracy. Synchronization guarantees that outputs from accelerators are exactly the same as we expected at all timesteps, which is referred to as time accuracy. Hence, we can eliminate synchronization when approximation is endurable. Minitaur~\cite{neil2014minitaur} and PEASE~\cite{roy2017programmable} adopt a totally event-driven architecture, where the neuron states updating only occurs at income spike arrivals. No synchronization is needed, but the length of timesteps becomes a variant, leading to inaccuracy compared with numerical computations in Algorithm~\ref{alg:1}. NeuroEngine~\cite{lee2021neuroengine} uses a bi-exponential model to reduce the complexity of solving the differential equation and add synapse delay support for SNN accelerators. 
However, an input spike can have effects on the next several timesteps when the input current ($I$ in Equation~\ref{equ:1}) decays over timesteps. In this case, updating neurons only at spike arrivals may make neurons generate spikes from the past timesteps, which makes inconsistent results with Algorithm~\ref{alg:1}. Sometimes we expect the hardware accelerators to precisely simulate SNNs we trained before, which limits applications of event-driven neuromorphic chips.

\begin{figure}
    \centering
    \subfloat[Workflow of SE.]{
        \hspace*{-1.5em}
        \includegraphics[width=0.54\linewidth]{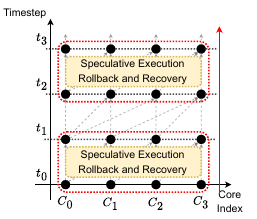}
        \label{fig:neurosync-workflow}
    }
    \subfloat[RR overheads.]{
        \includegraphics[width=0.45\linewidth]{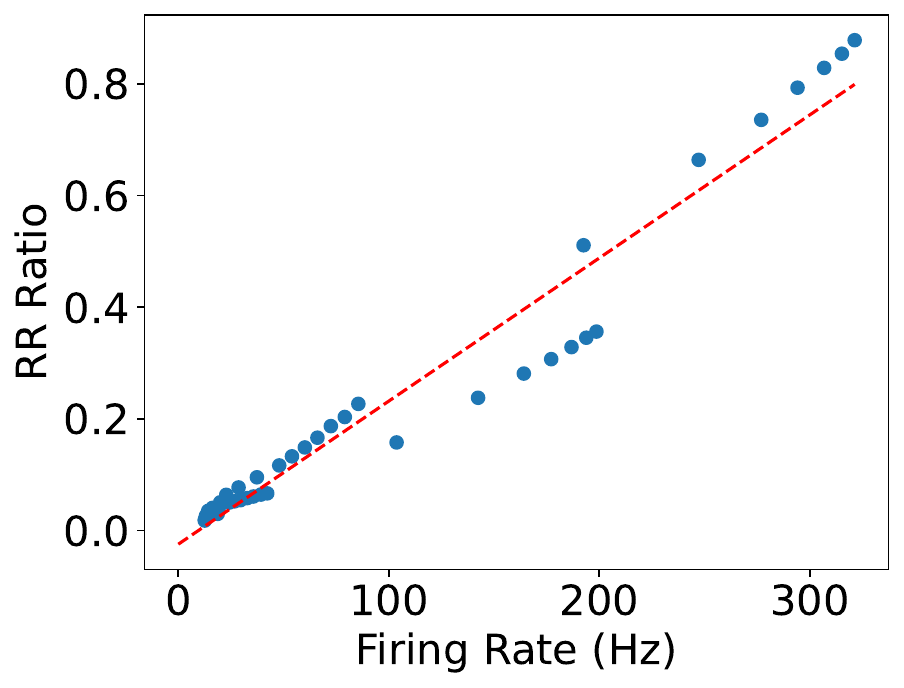}
        \label{fig:rr-overhead}
    }
    \caption{Synchronous SNN accelerators with speculative execution.}
    \vspace{-1.2em}
\end{figure}

Thirdly, incorporate additional mechanisms to mitigate synchronization. NeuroSync~\cite{lee2022neurosync} proposes a speculative-execution (SE) and rollback-recovery (RR) mechanism to reduce synchronization frequency. However, periodical all-core synchronization is still inevitable, and the rollback-and-recovery overheads grow with the firing rate increases. Fig.~\ref{fig:neurosync-workflow} gives a 4-core example of the SE mechanism: cores run speculatively with rollback and recovery until the period synchronization. Fig.~\ref{fig:rr-overhead} shows RR cycles of time-driven accelerators with speculative execution, identifying RR overheads become gradually inevitable when more spikes are generated. Speculative execution mechanisms cannot fully leverage their advantages without the cooperation of predictors. SpikeNC~\cite{xie2023spikenc} proposes a software agent-based asynchronous scheme. However, it does not leverage core dependencies and is limited at the SNN layer level. It also lacks hardware design. In this work, a novel time-accurate architecture is proposed to eliminate synchronization. We consider inter-cores dependencies as an additional control mechanism to achieve our design goals. Then, we realize DepAsync by adding a scheduler into each core and extending the spike communication mechanism.

It's worth noting that at circuit level, many-core neuromorphic chips often adopt a global asynchronous locally synchronous (GALS) manner~\cite{chapiro1984globally} in circuit design~\cite{akopyan2015truenorth,davies2018loihi,ma2017darwin}, which means the whole chip doesn't operate under single clock signal. The term 'synchronization' in this work is at the SNN computation level.

\section{DepAsync Design}
\subsection{Ignored Inter-Core Dependencies}
Line 6-9 in algorithm~\ref{alg:1} demonstrates that data dependencies between neurons are connected by synapses. At timestep $t$, post-synapse neurons update their internal states requiring spikes generated by pre-synapse neurons at timestep $t-1$. When we dispatch neurons to multicore SNN accelerators, dependencies between neurons are inherited by core dependencies. Similarly, we call core $A$ a \textit{pre-dependency} of core $B$ if $A$ sends spikes to $B$; versus the \textit{post-dependency}.

Noting that these dependencies are Read-After-Writes (RAWs), synchronization therefore becomes necessary in time-driven architectures for result correctness. In most SNN accelerators, synchronization is rigid, since it requires all cores to wait together, like a barrier primitive. However, each core is supposed to wait for only part of the cores which it depends. When all pre-dependencies of a core finish timestep $t$ and all spikes are received, it can safely forward to the next timestep without waiting for other unfinished cores.

\begin{figure}
    \centering
    \subfloat[Topology.]{
        \raisebox{.35\height}{
            \includegraphics[width=0.2\linewidth]{figures/topology.pdf}
        }
    }
    \subfloat[Workflow of DepAsync.]{
        \includegraphics[width=0.45\linewidth]{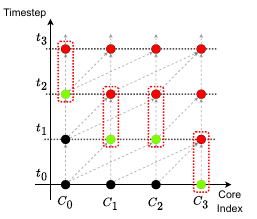}
        \label{fig:depasync-workflow}
    }
    \caption{Dependencies diagram of DepAsync.}
    \label{fig:depasync}
    \vspace{-1.2em}
\end{figure}

Fig.~\ref{fig:depasync} illustrates a 4-core example, where grey arrows denote data dependencies same as the topology at the left, meanwhile, dots in black, green, and red represent core workloads at finished, running, and unfinished timesteps, respectively. At this moment, core 0 is running at timestep $t_2$, core 1-2 are at $t_1$, and core 3 is at $t_0$. For core 0, no pre-dependency means it can keep forwarding to the next, while core 1 and 2 rely on spikes from core 0, thus they can only safely proceed $t_2$ after the current timestep finishes without waiting for the slowest core 3. Finally, core 3 only has a forward window $[t_0, t_1]$, since both core 1 and 2 are its pre-dependencies.

On the other hand, post-dependencies limit the size of the \textit{forward window} (red rectangles in Fig.~\ref{fig:depasync-workflow}), which means how many timesteps a core can forward without waiting for its post-dependencies. In neuromorphic cores, the memory size assigned for buffering input spikes is finite, which means there may be no more empty buffer for spikes generated at far future timesteps. Here, we set the size of the forward window is 2 for visualization. For example, with constraints of core 1 and 2 to buffering input spikes from 2 more timesteps, the forward window of core 0 will be $[t_2, t_3]$. In this case, if core 0 runs over $t_3$ fast enough that core 1 or 2 is still working on $t_1$, it has to wait for them to release some spike buffers when starting $t_2$.

\subsection{Tracing Dependencies and Running States}
The dependency analysis above gives neuromorphic cores the ability to get rid of whole-system synchronization. To implement such a mechanism, the main challenge is that each core has to trace dependencies and their running states.

\textbf{Storing dependencies.} For most SNNs, once the network is defined, connections (not weights) between neurons are determined. When deployed to accelerators, SNNs are first partitioned into several groups, each of which corresponds to a logic neuromorphic core. Then these logic cores are mapped to physical cores in real hardware by mapping algorithms~\cite{jin2023mapping} designed for reducing spike transfer latency and on-chip network communication congestion. After mapping, fixed real core dependencies can be stored at in-core memories as other neuron parameters.

\textbf{Tracing running states.} Unlike static dependencies, running states dynamically change during whole the SNN inference, which is traced by NoCs in our proposed architecture. NoCs in most SNN accelerators are mainly used for transferring generated spikes. As the communication component of the system, they are also responsible for transferring meta information, such as input parameters~\cite{ma2017darwin}, core statistics~\cite{furber2014spinnaker}, and synchronization messages~\cite{davies2018loihi}. Naturally, we allow neuromorphic cores to transmit running states to their dependencies.

\begin{figure}[t]
    \centering
    \includegraphics[width=\linewidth]{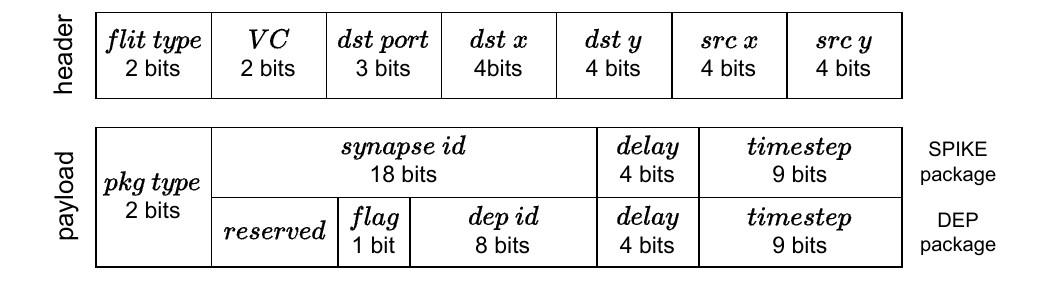}
    \caption{Package structure in NoC. $VC$: out virtual channel; $dst\ port$: next hop destination; $dst/src\ x/y$: destination/source position; $synapse\ id$: activated synapse id in post-dependencies; $delay$: synapse delay; $timestep$: spike timestep; $flag$: START/FINISH flag; $dep\ id$: dependency id.}
    \label{fig:package}
    \vspace{-1.2em}
\end{figure}

Generally, packages in NoCs compose package headers including meta information like package type and routing information, and package bodies containing different kinds of message payloads. As shown in Fig.~\ref{fig:package}, we add a new kind of package, DEP, for transmitting running states. The DEP package body has three components. The \textit{identifier} entry denotes the dependency the package belongs to. The \textit{timestep} entry records the running progress of dependencies. The \textit{flag} entry is for distinguishing running states. In this work, \textit{flag} is only 1 bit, with 0 for a timestep finish (FINISH package), and 1 for a timestep start (START package). Then, the core procedure in DepAsync is changed to Algorithm~\ref{alg:2}, where red lines highlight the differences.

In each neuromorphic core, two tables are prepared for running states of dependencies, which will be updated at the arrival of START/FINISH packages. Assume a core can accept spikes from $m$ future timesteps, its safe forwarding conditions at timestep $t$ is two-fold: 1) all its pre-dependencies finish $t$; 2) all its post-dependencies start $t - m + 2$. If any condition is unsatisfied, the core turns to a wait state and keeps receiving and monitoring new packages. Once new DEP packages arrive and carry relevant information, the core is triggered to forward to the next timestep. When $m = 1$, cores holding the output layer of the SNN, which have no post-dependencies, will first forward to $t+1$, then wake up other cores in a cascading way. In this case, the whole system falls back to the synchronization mechanism. Otherwise, if $m > 1$, DepAsync can provide fine-grained control of neuromorphic cores with output correctness at every timestep.

\begin{algorithm}[b]
    \renewcommand{\algorithmicrequire}{\textbf{Input:}}
    \renewcommand{\algorithmicensure}{\textbf{Output:}}
    \caption{Neuromorphic core procedure in DepAsync}
    \label{alg:2}
    \begin{algorithmic}[1]
        \WHILE{$t < t_{max}$}
            \color{red}
            \IF{not allowed to forward to $t$}
                \STATE waiting
            \ENDIF
            \STATE send START packages to its pre-dependencies
            \color{black}
            
            \STATE calculate weight sum, apply learning rules
            \FORALL{$x_i$ in this core}
                \STATE update $x_i.states$
                \IF{$x_i$ fires a spike}
                    \STATE send spike packages to post-dependencies
                \ENDIF
            \ENDFOR
            
            \color{red}
            \sout{waiting synchronization}
            \STATE send FINISH packages to its post-dependencies
            \color{black}
            
            \STATE $t = t + 1$
        \ENDWHILE
    \end{algorithmic}  
\end{algorithm}

\begin{figure*}[ht]
    \centering
    \includegraphics[width=\linewidth]{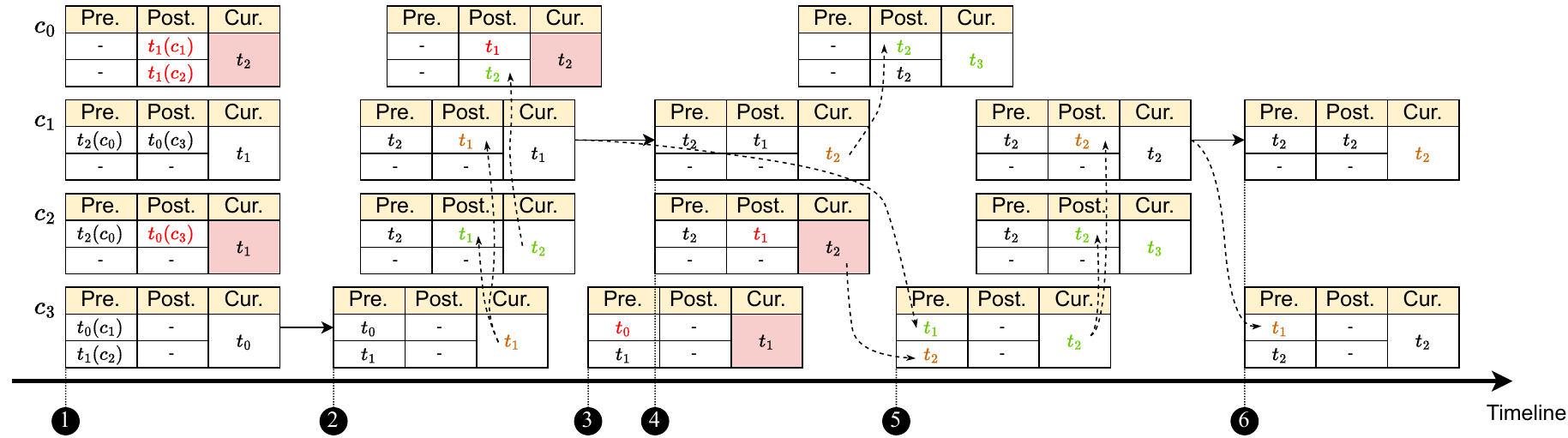}
    \caption{A 4-core example of the DepAsync workflow with two spike buffers.}
    \label{fig:DepAsync-example}
    \vspace{-1em}
\end{figure*}

Formally, the safe forwarding condition at Line 2 in Algorithm~\ref{alg:2} is calculated by Equation~\ref{equ:2}:
\begin{equation}
    \label{equ:2}
    \begin{aligned}
        \text{cond}
            &= \text{pre-cond} \wedge \text{post-cond} \\
            &= (\bigwedge_{i=1}^{N_{pre}} t_{i}^{pre} >= t^{cur}) \wedge (\bigwedge_{i=1}^{N_{post}} t_{i}^{post} > t^{cur} - m + 1)
    \end{aligned}
\end{equation}
where $N_{pre}, N_{post}$ are the number of pre- and post-dependencies, $t_{i}^{pre},t_{i}^{post}$ denote the timestep from received FINISH/START packages, and $t^{cur}$ represents the current timestep of each neuromorphic core.

Fig.~\ref{fig:DepAsync-example} gives an example of the DepAsync workflow. The topology is the same as Fig.~\ref{fig:depasync}, and the spike buffer can hold spikes from 2 timesteps (i.e., $m=2$). Table headers \textit{pre}, \textit{post} and \textit{cur} denote tables for pre-dependencies, post-dependencies, and the current timestep, respectively. The $c_i$ within the parentheses at the beginning represents core dependency topology corresponding to Fig.~\ref{fig:depasync}. Solid arrows represent safe forwarding and dash arrows are DEP packages sent by neuromorphic cores. At the start moment \ding{182}, we also assume running states of cores align to Fig.~\ref{fig:depasync} and timestep $t_3$ is the end. In detail, core $c_0$ has finished timestep $t_2$, $c_1$ is running at $t_1$, while $c_2$ is at the end of $t_1$, finally, $c_3$ is still working at $t_0$. In this case, contents in tables for pre- and post-dependencies are determined. $c_0$ is stopped after $t_2$ (filled in the red cell) because of the post-condition (illustrated as a red font), and $c_2$ is stuck at $t_1$ for the same reason. At moment \ding{183}, $c_3$ safely forwards to its next timestep $t_1$. Since $c_1$ and $c_2$ are pre-dependencies of $c_3$, START packages at $t_1$ from $c_3$ are generated and sent to them. After a transmit latency in the NoC, $c_1$ and $c_2$ receive the packages and update their post-tables. In particular, the update in $c_2$ satisfies the post-condition, thus $c_2$ is wakened up to work on $t_2$, which also sends a START@2 package to $c_0$. When it comes to moment \ding{184}, $c_3$ surpasses the progress of $c_1$, which means it has to wait for pre-dependency $c_1$ to finish $t_1$ first. Then at moment \ding{185}, $c_1$ accomplishes $t_1$ and successfully goes to the next timestep, therefore it sends a FINISH@1 to $c_3$ and a START@2 to $c_0$. Simultaneously, $c_2$ ends $t_2$ and waits $c_3$ again. The FINISH@1 from $c_1$ and FINISH@2 from $c_2$ arrive $c_3$ at time \ding{186}, then replace the timesteps recorded in the pre-table and trigger $c_3$ to proceed to $t_2$. Then $c_3$ triggers $c_2$ after a while. Finally at moment \ding{187}, $c_3$ goes to the last timestep, sending FINISH@2, which satisfies pre-cond of $c_3$. Overall, neuromorphic cores in DepAsync alternately proceed until the simulation end, and leverage dependencies to ensure correct results without all-core synchronization.

\section{DepAsync Implementation}
\begin{figure}
    \centering
    \includegraphics[width=\linewidth]{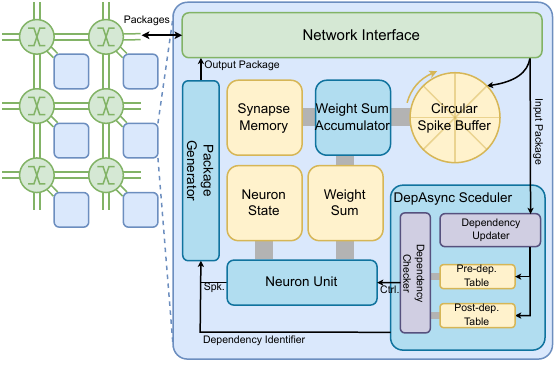}
    \caption{Overview of DepAsync architecture}
    \label{fig:arch}
    \vspace{-1em}
\end{figure}

In this section, we implement DepAsync by three key components in neuromorphic cores. Fig.~\ref{fig:arch} gives an overview of our architecture. We first introduce the novel DepAysnc Scheduler, which plays a core role in dependency analysis. Then, we design our spike buffers used for storing future spikes, making it possible to proceed neuromorphic core without whole-system waiting. Next, we enhance the package generator to support new DEP packages cooperating with the DepAysnc scheduler. Finally, routers in NoCs are modified, ensuring correctness during the DEP package transmission. It is worth noting that the DepAsync mechanism is suitable for a wide range of neuron and synapse models, and we adopt the computation (including neuron update and synapse accumulation) and firing control components similar to current accelerators~\cite{davies2018loihi,lee2022parallel}. For example, the LIF model needs a multiplier to implement the leak mechanism, an adder to add on the accumulation that is accumulated after spike arrivals, and a comparator to detect whether the membrane voltage exceeds the threshold (i.e. firing a spike). 

\subsection{DepAsync Scheduler}

\begin{figure}
    \centering
    \includegraphics[width=\linewidth]{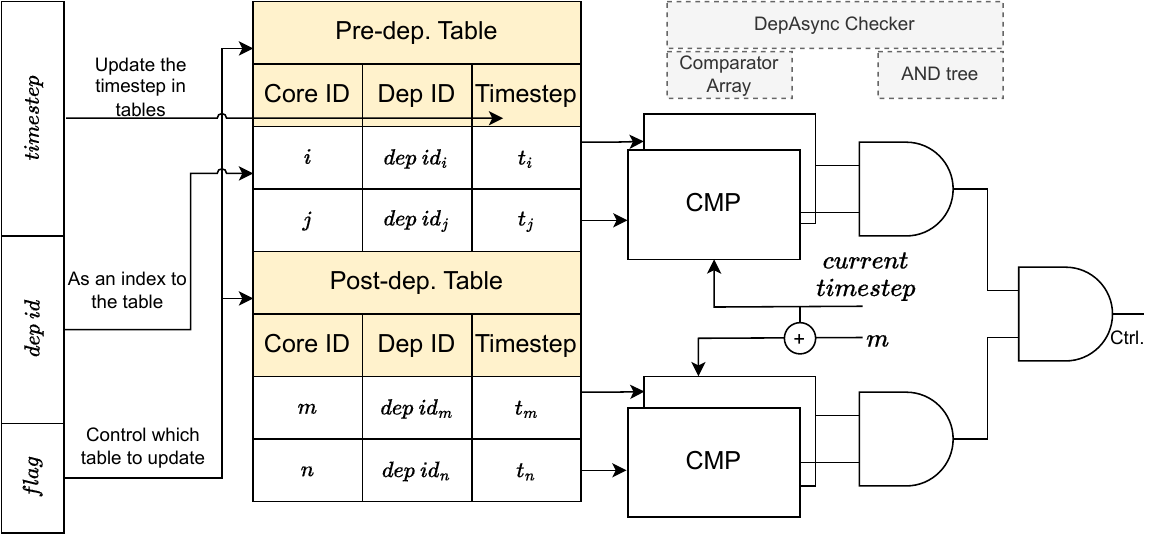}
    \caption{Datapath in the DepAsync scheduler.}
    \label{fig:scheduler}
    \vspace{-1em}
\end{figure}

The main component for core dependency is the DepAsync Scheduler with a working mechanism illustrated as Fig.~\ref{fig:scheduler}. The scheduler is responsible for tracing dependency running states and controlling core forwarding.

\textbf{Tracing dependency}. The scheduler keeps monitoring income packages from the network interface. When DEP packages arrive, the scheduler will store the carried $timestep$ into the pre- or post-dependency table decided by the $flag$ entry, with the address given by $dep\ id$.

\textbf{Controlling.} On the other hand, the scheduler keeps generating control signals to direct the neuromorphic core whether forwarding to next. The timesteps stored in the pre-dependency table are compared with the current timestep $t$, checking whether there are potential unarrived spikes required by the next timestep. Meanwhile, all timesteps stored in the post-dependency table are compared with $t - m$, ensuring all post-dependencies have enough space for buffering spikes generated at $t + 1$. These comparators and AND gates implement Equation~\ref{equ:2}. Once a timestep is finished or a DEP package arrives, the scheduler will update the related entry in the dependency table and update the control signal, as demonstrated in Algorithm~\ref{alg:2}. 


\subsection{Circular Spike Buffer}
The spike buffer in DepAsync is circular, storing incoming spikes for subsequent processing. At the end of each timestep, buffers are rotated one slot. Circular spike buffers are generally adopted for supporting synapse delay~\cite{rhodes2018spynnaker}, which means spikes received at $t$ are used at $t+delay$. In this work, other slots store future spikes from pre-dependencies, logically the same as spikes with some delay. DepAsync works with synapse delay as well, subject to:

\begin{equation}
    \label{equ:3}
    N_{slot} = max\_delay + m - 1
\end{equation}

\begin{figure}
    \centering
    \includegraphics[width=0.6\linewidth]{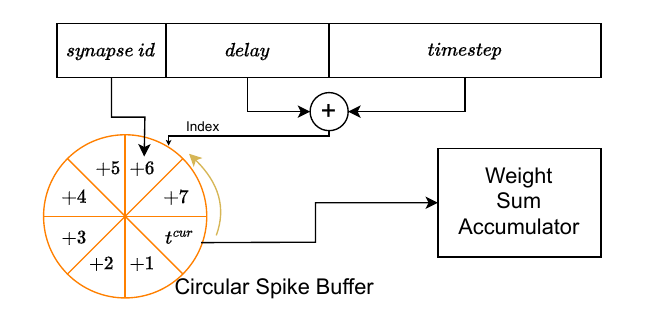}
    \caption{Datapath of the circular spike buffer.}
    \label{fig:spike-package}
    \vspace{-1em}
\end{figure}

Fig.~\ref{fig:spike-package} shows how to store a spike in the circular buffer. The $timestep$ entry added by the $delay$ determines which slot to access. As constrained by the scheduler, the slot destination never conflicts with the slot being used.

\subsection{Package Generator}
To work with dependency analysis performed by the scheduler, the package generator is supposed to correctly send packages of DEP type. 

\begin{figure}[t]
    \centering
    \includegraphics[width=0.8\linewidth]{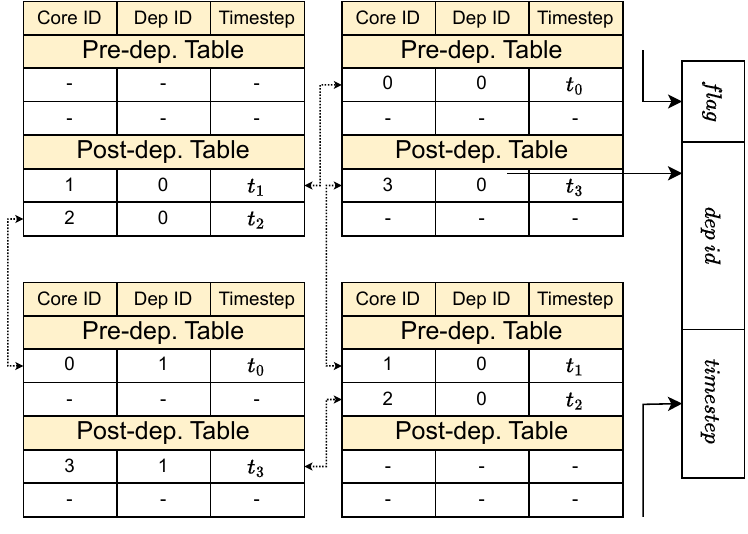}
    \caption{Datapath for generating DEP packages.}
    \label{fig:package-generator}
    \vspace{-0.3em}
\end{figure}

As illustrated in Fig.~\ref{fig:package-generator}, entry $flag$ and $timestep$ in the DEP package are trivial, directly determined by the DEP package type and the current timestep. On the other hand, we need an additional mechanism to generate the correct $dep\ id$. As mentioned before, $dep\ id$ is for addressing dependency tables. Thus, a neuromorphic core ought to be aware of its reversed index in pre- and post-dependencies, which is static and determined along with dependency relationships. In DepAsync, we store them at dependency tables. Fig.~\ref{fig:package-generator} gives an example corresponding to the topology in Fig.~\ref{fig:depasync}. The entry DEP ID stored in one core points to the index in its dependencies. Then the destination core of a DEP package can decide which line in the dependency table to write.

\subsection{Multi-VC NoC Router}
The basic assumption in DepAsync is that START/FINISH packages precisely represent the start and finish of a timestep of neuromorphic cores, which implies that the START/FINISH packages must arrive at destination cores in order, and the FINISH package at timestep $t$ must be later than all spike packages. It is naturally satisfied if the on-chip network orderly transfers packages.


Recently, prevalent NoCs adopt the virtual channel technique to improve network performance. Each router port has more than one physical channel and concurrently transmits several packages by multiple channels. Each input port in a router has an arbiter to grant a request from multiple virtual channels at each cycle in a round-robin strategy~\cite{monemi2017pronoc}, which makes the package transmission out of the original order and violates the assumption in DepAsync. 

\begin{figure}
    \centering
    \includegraphics[width=0.8\linewidth]{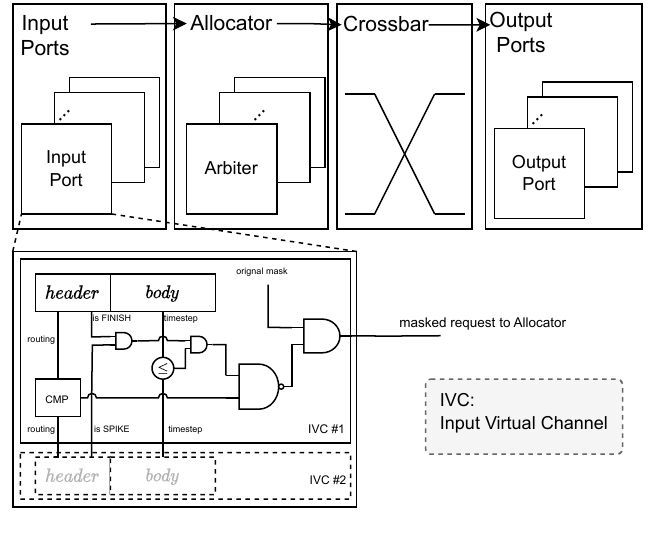}
    \caption{Datapath of the additional mask in input virtual channels.}
    \label{fig:noc}
    \vspace{-1em}
\end{figure}

To cooperate DepAsync with virtual channels, the spike packages should be prior to the FINISH packages with the same source, destination, and timestep. Fortunately, routers with virtual channels generally mask out invalid channels before arbitrating one to transmit~\cite{monemi2017pronoc}. Thus, an additional mask is attached as shown in Fig.~\ref{fig:noc}. In this case, FINISH packages always wait for spikes transfer first, and START packages are orderly transmitted as well. Noticing that the out-order of spikes within the same timestep is insignificant for correctness, DepAsync is capable of benefiting from performance improvement due to virtual channels.

\section{Evaluation}
\subsection{Experimental Setup}

\subsubsection{Experimental Platform} \label{sec:eva}
To model performance, we develop a cycle-accurate simulator to obtain workload latency. Neuromorphic cores execute instructions in Algorithm~\ref{alg:2} cycle by cycle, and the NoC is modified based on HeteroGarnet~\cite{heterogarnet}. Table~\ref{tab:simulator} lists part of the parameters.

\begin{table}[h]
    \centering
    \caption{DepAsync parameters}
    \vspace{-0.7em}
    \begin{tabular}{c|c}
        \hline
        \textbf{Parameter} & \textbf{DepAsync simulator} \\ \hline \hline
        Topology & 2D mesh \\ \hline
        Grid size & \makecell{configurable \\ (eg. $4\times4$, $16\times8$)} \\ \hline
        Cycles per NoC Hop & 2 cycles \\ \hline
        NoC virtual channels & 4 \\ \hline
        Routing Algorithm & XY routing \\ \hline
    \end{tabular}
    \label{tab:simulator}
    \vspace{-0.5em}
\end{table}

The energy model used in this work originates from Eyeriss~\cite{chen2016eyeriss}, where the energy cost of a single operation (computation or data movement) is first synthesized and modeled, and then the overall energy consumption is estimated by combining it with the number of operations. This model has been validated~\cite{parashar2019timeloop,wu2019accelergy}, and become prevalent SNN accelerators~\cite{narayanan2020spinalflow,yin2022sata,lee2022parallel,liu2022sato}. In this work, DepAsync is evaluated at 400MHz using 65nm CMOS technology, with memory units estimated by CACTI~\cite{balasubramonian2017cacti}. DepAsync uses 270.3 KB and 8 KB SRAM for each core to hold synapses and neurons, and an additional 4 KB for the scheduler to hold 512 core dependencies. The limitation of the number of dependencies leads to a new constraint in compilation. The spike buffer is dependent on the parameter $m$, consuming $4m$ KB SRAM. In the following experiments, $m$ is 4 unless otherwise specified. Then we use a 2-cycle router with 4 virtual channels for communication. In a 64-core chip with $m=4$ and 4 virtual channels, the area of each component is listed in Table~\ref{tab:area}. The area overhead caused by DepAsync is 6.76\%.

\begin{table}[h]
    \centering
    \scriptsize
    \caption{Chip Area ($mm^2$)}
    \vspace{-0.7em}
    \begin{tabular}{c|c|c}
        \hline
        \textbf{Component} & Sync & DepAsync \\ \hline \hline
        Neuron & 7.71 & 7.71  \\
        Synapse & 228.20 & 228.20 \\
        Communication & 21.60 & 22.61 \\
        Spike Buffer & 3.77 & 16.51 \\
        Scheduler & - & 7.95 \\ \hline
        Total & 265.07 & 282.99 \\ \hline
    \end{tabular}
    \label{tab:area}
    \vspace{-0.5em}
\end{table}

Overall, the evaluation pipeline for DepAysnc is as follows. Firstly, the SNN workload and configuration generated by compilers are first fed into the cycle-accuracy simulator. Then, the simulator simulates the SNN and output statistics such as the number of neuron updating operations, memory access, and NoC hops. Finally, the analytic energy model is invoked to generate overall energy consumption.


\subsubsection{Baselines}
We compare DepAsync with two time-driven baselines, Loihi~\cite{davies2018loihi} and NeuroSync~\cite{lee2022neurosync}. As Loihi is a commercial SNN accelerator, we carefully implement the synchronization mechanism of Loihi in our simulator (referred to as \textbf{Sync}) for convenience. Simulator parameters in Table~\ref{tab:simulator} are all available for the Sync baseline mode. On the other hand, the speculative-execution and rollback-and-recovery mechanisms in NeuroSync are developed as well (referred to as \textbf{SE}). It is worth noting that NeuroSync has proposed a Defer Learning method to reduce rollbacks caused by post-learning, which is however not useful to compare with DepAsync because misspeculation caused by RAWs in this work cannot be avoided. We also validate the output spike results of the two baselines to guarantee the exact time accuracy same as DepAsync.

We do not compare against time-driven accelerators with fixed synchronization signals such as TrueNorth~\cite{akopyan2015truenorth} and SpiNNaker~\cite{furber2014spinnaker}, because their synchronization frequency relies on user settings. It is inconvenient to choose a proper frequency for all SNN workloads. Loihi provided a more adaptive way to implement synchronization.

\subsubsection{Workloads} \label{sec:benchmark}
We select five different SNN workloads to evaluate DepAsync. The workloads have different SNN structures and scales, as listed in Table~\ref{tab:benchmark}.

\begin{table}[h]
    \centering
    \scriptsize
    \caption{Evaluation workloads. \textbf{I/O}: ifmap/ofmap width/height; \textbf{IC/OC}: \# of ifmap/ofmap channels; \textbf{K}: kernel width/height; \textbf{S}: kernel stride. }
    \vspace{-0.7em}
    \begin{tabular}{c|c|c|c|c|c|c|c}
        \hline
        \textbf{Workload} & \textbf{Layer} & \textbf{I} & \textbf{IC} & \textbf{K} & \textbf{S} & \textbf{O} & \textbf{OC} \\ \hline \hline
        \multirow{5}{*}{\makecell{MNIST\\Timestep: 500\\$4\times4$ cores}}
        & Conv1 & 28 & 1 & 5 & 2 & 12 & 16 \\
        & Conv2 & 12 & 16 & 3 & 1 & 10 & 32 \\
        & Pool1 & 10 & 32 & 2 & 1 & 5 & 32 \\
        & Conv3 & 5 & 32 & 3 & 1 & 5 & 8 \\
        & FC1 & 5 & 8 & 8 & 1 & 1 & 10 \\
        \hline

        \multirow{5}{*}{\makecell{NMNIST\\Timestep: 500\\$4\times4$ cores}}
        & Conv1 & 34 & 1 & 5 & 1 & 30 & 16 \\
        & Pool1 & 30 & 16 & 2 & 1 & 15 & 16 \\
        & Conv2 & 15 & 16 & 3 & 1 & 13 & 32 \\
        & Pool2 & 13 & 32 & 2 & 1 & 6 & 32 \\
        & FC1 & 6 & 32 & 6 & 1 & 1 & 10 \\
        \hline

        \multirow{5}{*}{\makecell{DVSGesture\\Timestep: 500\\$8\times8$ cores}}
        & Conv1 & 128 & 1 & 5 & 2 & 62 & 16 \\
        & Pool1 & 62 & 16 & 2 & 1 & 31 & 16 \\
        & Conv2 & 31 & 16 & 5 & 2 & 14 & 32 \\
        & Pool2 & 14 & 32 & 2 & 1 & 7 & 32 \\
        & FC1 & 7 & 32 & 7 & 1 & 1 & 11 \\
        \hline

        \multirow{5}{*}{\makecell{CIFAR10DVS\\Timestep: 500\\$8\times8$ cores}}
        & Conv1 & 128 & 1 & 5 & 2 & 62 & 32 \\
        & Pool1 & 62 & 32 & 2 & 1 & 31 & 32 \\
        & Conv2 & 31 & 32 & 5 & 2 & 14 & 64 \\
        & Pool2 & 14 & 64 & 2 & 1 & 7 & 64 \\
        & Conv3 & 7 & 64 & 3 & 1 & 5 & 128 \\
        & FC1 & 5 & 128 & 5 & 1 & 1 & 10 \\
        \hline
    \end{tabular}
    \label{tab:benchmark}
\end{table}

The synthetic workload is generated like Izhikevich et al.~\cite{izhikevich2003simple}, with both excitatory and inhibitory neurons involved. Synapses and connection weights are randomly set. We choose this workload for a wide range of workload scales and neuron parameters(Table~\ref{tab:scalability}), which are useful to evaluate the scalability of the proposed DepAsync. Then, we utilize two workloads with real data. The SNN for MNIST~\cite{lecun2010mnist} is converted from a pre-trained ANN by snntoolbox~\cite{rueckauer2017conversion}, and dense images from the dataset are encoded by a Poisson process. On the other hand, the NMNIST~\cite{orchard2015converting}, CIFAR10DVS~\cite{li2017cifar10dvs} and DVS-Gesture dataset~\cite{amir2017low} contain real spike trains generated from digital images, CIFAR10 images and hand gestures recorded by dynamic vision sensor (DVS) cameras, which are used to train an SNN by the STCA learning rule~\cite{gu2019stca} in a SNN framework SPAIC~\cite{hong2022spaic}. It is worth noting that our DepAsync focuses on the synchronization issue and is suitable for step-by-step simulated SNNs with different neuron and synapse models. In our experiments, we use the LIF model in the above workloads.

We convert different workloads to a common configuration format containing SNN structures and sample data. The SNNs are partitioned layer-by-layer where each layer takes several logic cores. Then we adopt an efficient mapping algorithm~\cite{jin2023mapping} to map high-level SNN logic cores to low-level hardware neuromorphic cores for future simulation.

\subsection{Overall Results}

\begin{figure}
    \centering
    \subfloat[Speedup.]{
        \hspace{-1.5em}
        \includegraphics[width=0.5\linewidth]{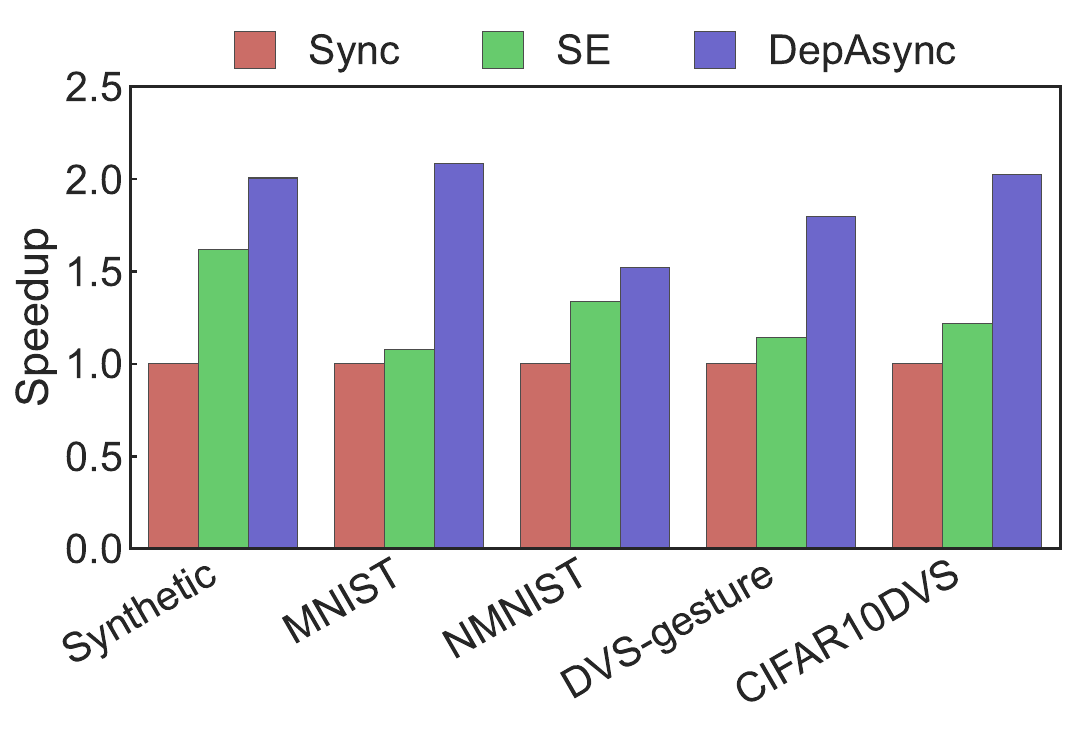}
        \label{fig:speedup}
    }
    \subfloat[Energy efficiency.]{
        \includegraphics[width=0.5\linewidth]{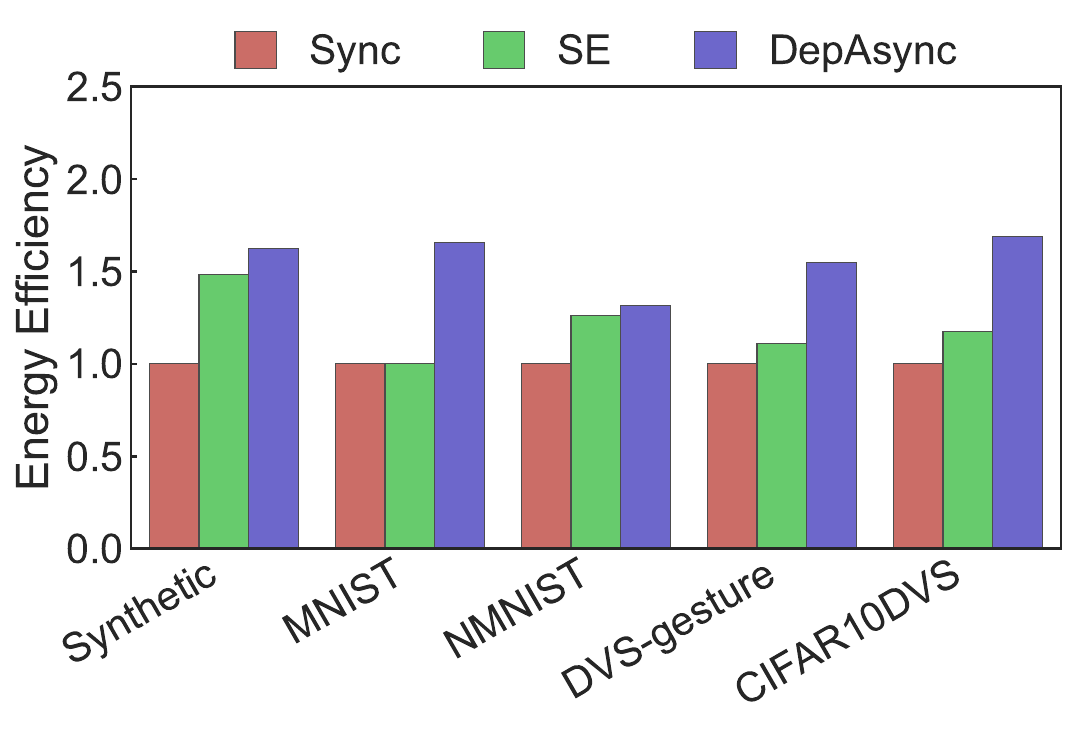}
        \label{fig:energy}
    }
    \caption{Overall results.}
    \label{fig:performance}
    \vspace{-1.5em}
\end{figure}

\begin{figure}
    \centering
    \subfloat[Latency breakdown.]{
        \includegraphics[width=0.5\linewidth]{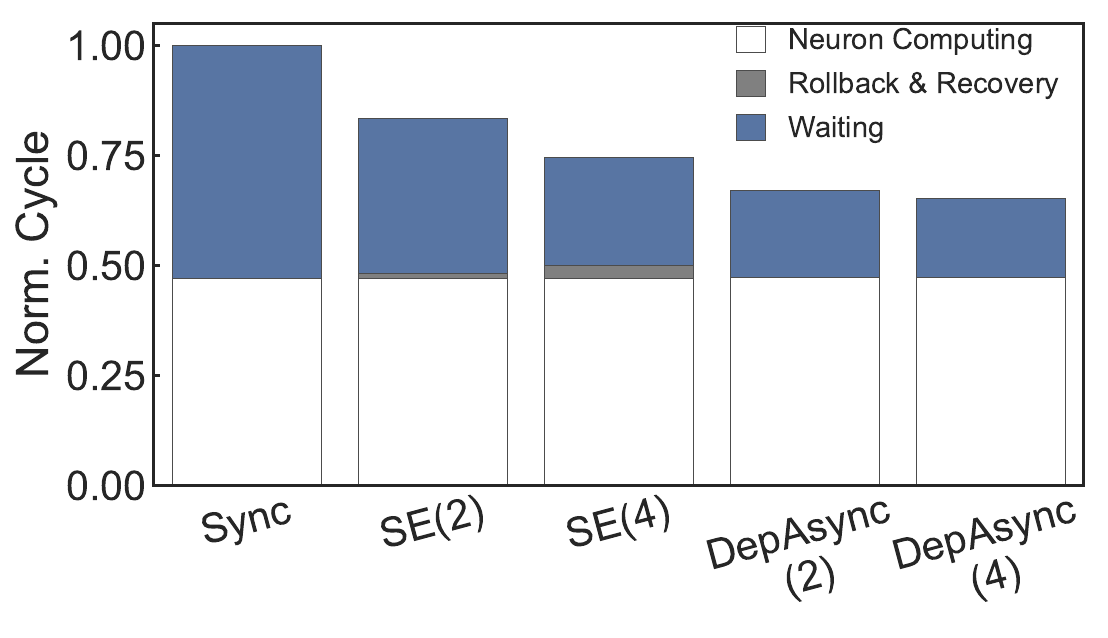}
        \label{fig:breakdown}
    }
    \subfloat[Energy breakdown.]{
        \includegraphics[width=0.4\linewidth]{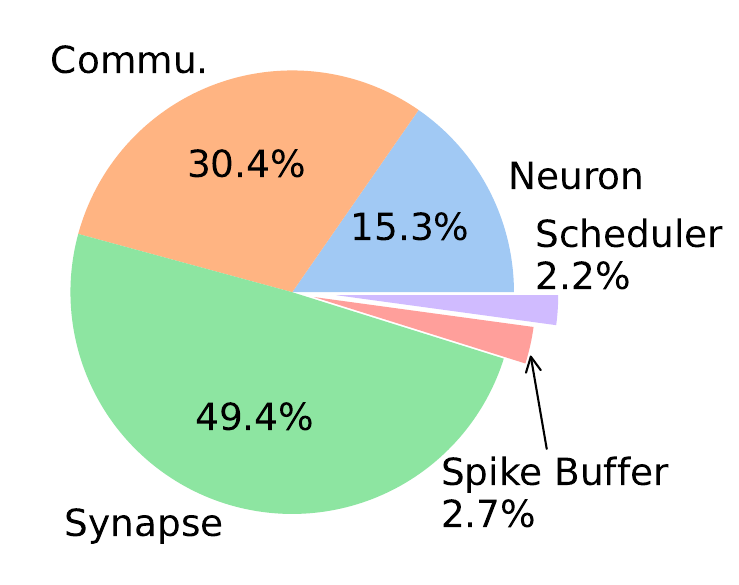}
        \label{fig:breakdown-energy}
    }
    \caption{Breakdown of DepAsync.}
    \vspace{-1.5em}
\end{figure}

Compared with the two baselines, we show the overall speedup and energy efficiency of DepAsync. Fig.~\ref{fig:performance} illustrates the improvement of speculative execution (SE) and DepAsync over synchronous accelerators (Sync) on five workloads. In summary, DepAsync achieves 1.86x and 1.55x harmonic mean speedup and energy efficiency over plain synchronous time-driven accelerators, meanwhile 1.46x, and 1.29x over speculative-execution architectures. Exploiting more core parallelism than Sync speeds up DepAsync. And DepAsync performs better than SE because it schedules cores by core dependencies, avoiding misspeculation and rollback overheads.

DepAsync leverages core dependencies to eliminate all-core synchronization. In the worst case, if a core with many post-dependencies is always the slowest one, these post-cores still wait at every timestep, which is formally equivalent to a many-core synchronization. However, this situation is quite unlikely to occur in a real data SNN workload.

\begin{figure}
    \centering
    \includegraphics[width=\linewidth]{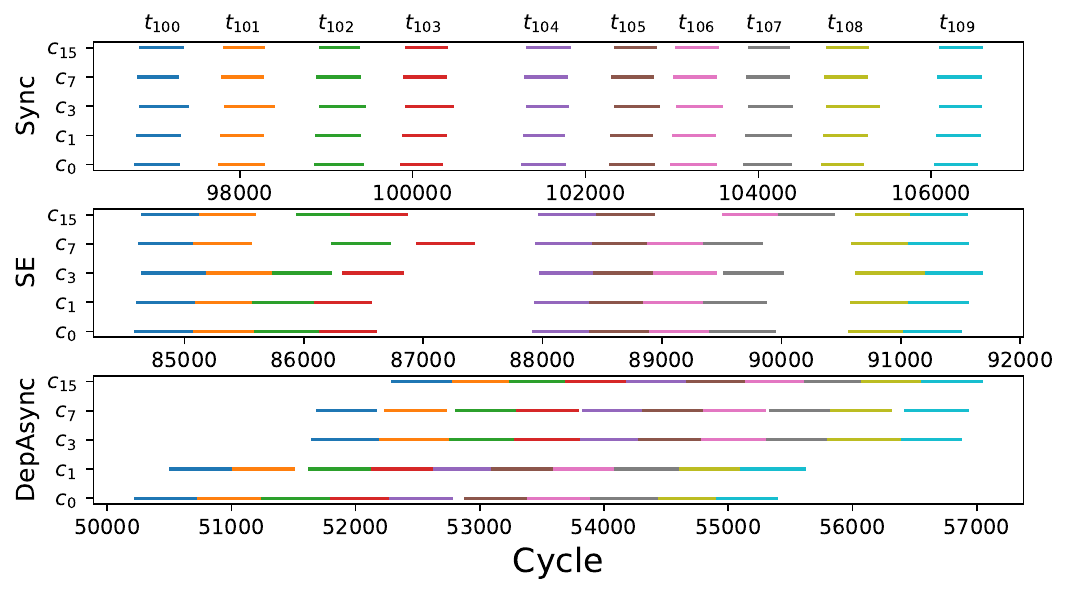}
    \caption{A slice when three architectures running from timestep 100 to timestep 109.}
    \label{fig:run}
    \vspace{-1.5em}
\end{figure}

Fig.~\ref{fig:breakdown} shows the latency breakdown of Sync, SE, and DepAsync. The waiting time in both SE and DepAsync decreases as the level of relaxation to synchronization increases. Although the idle time of SE cores is less than DepAsync, the cost of RR operations remains and even slightly grows due to the more aggressive speculative execution. For fair comparison between SE and DepAsync, the rest experiments set the synchronization period the same as the number of spike buffer slots $m$. Fig.~\ref{fig:breakdown-energy} shows the DepAsync scheduler and additional spike buffers take about 5\% of total energy consumption. 

Fig.~\ref{fig:run} captures a slice in the MNIST workload, showing the difference between DepAsync and synchronous architectures, where colored lines represent neuron computation in a single timestep. Periodic synchronization can be observed from blanks lying in two workload lines, meanwhile, DepAsync makes neuromorphic cores consecutively execute computations.

\subsection{Sensitivity Analysis}
\subsubsection{Spike Buffer Size}
\begin{figure}
    \centering
    \subfloat[Speedup.]{
        \hspace*{-1.5em}
        \includegraphics[width=0.5\linewidth]{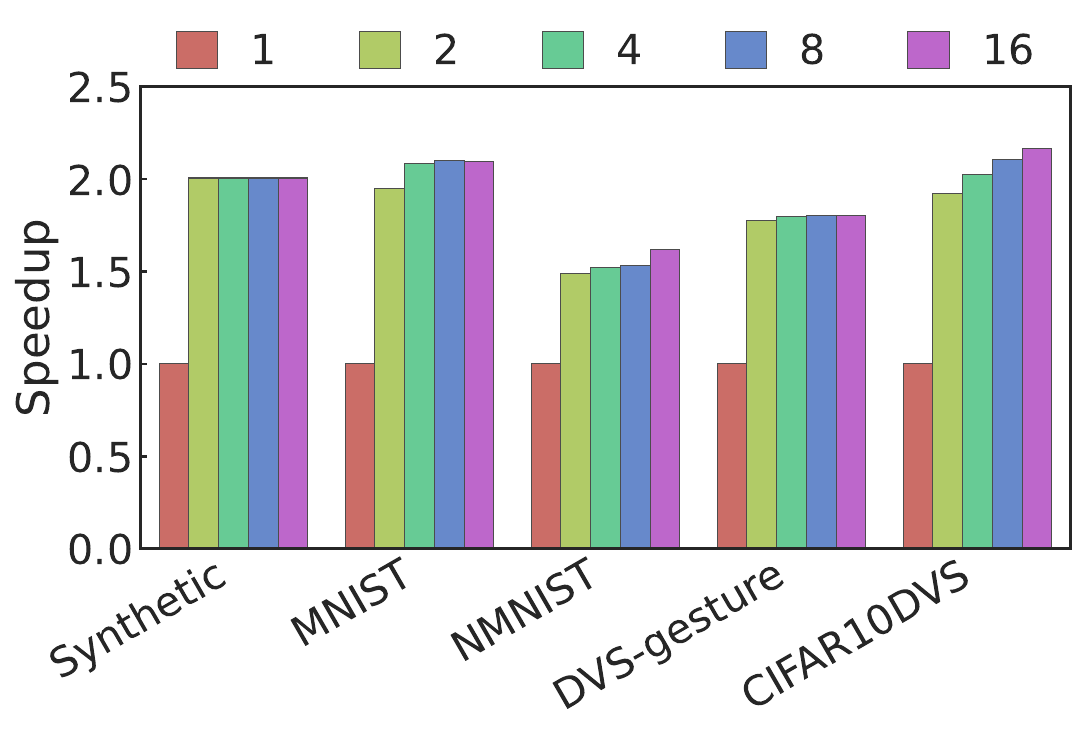}
        \label{fig:speedup-inb}
    }
    \subfloat[Energy efficiency.]{
        \includegraphics[width=0.5\linewidth]{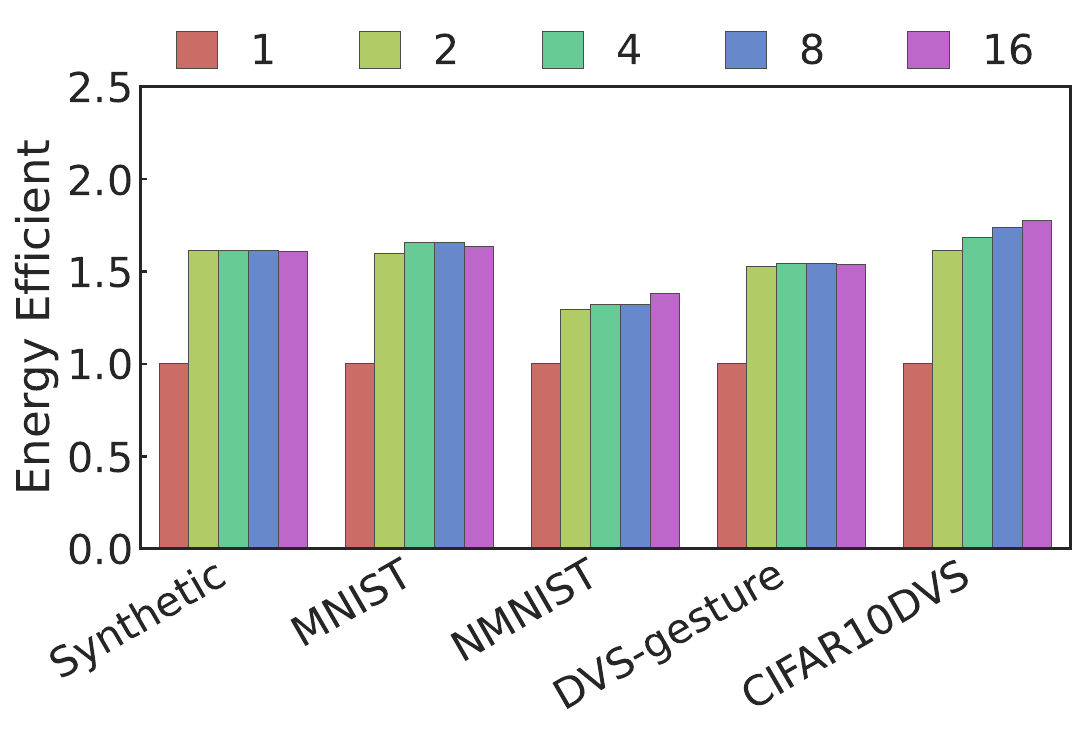}
        \label{fig:energy-inb}
    }
    \caption{The impact of the spike buffer size. The number in the legends denotes the number of spike buffer slots (i.e. $m$).}
    \label{fig:inb}
    \vspace{-1em}
\end{figure}

Fig.~\ref{fig:inb} shows the impact of the spike buffer size on DepAysnc performance. When the spike buffer size becomes large, more future timestep spikes can be stored, leading to a larger forwarding window and better performance. Different workloads have different performance converge speeds. Up to 16x spike buffer size, DepAsync gains 92\% and 58\% speedup and energy efficiency than baseline, respectively. Moreover, large spike buffers occupy more memory resources. Thus, it is a trade-off in real applications.

\subsubsection{Firing Rate}
\begin{figure}
    \centering
    \subfloat[Speedup.]{
        \hspace*{-1.5em}
        \includegraphics[width=0.5\linewidth]{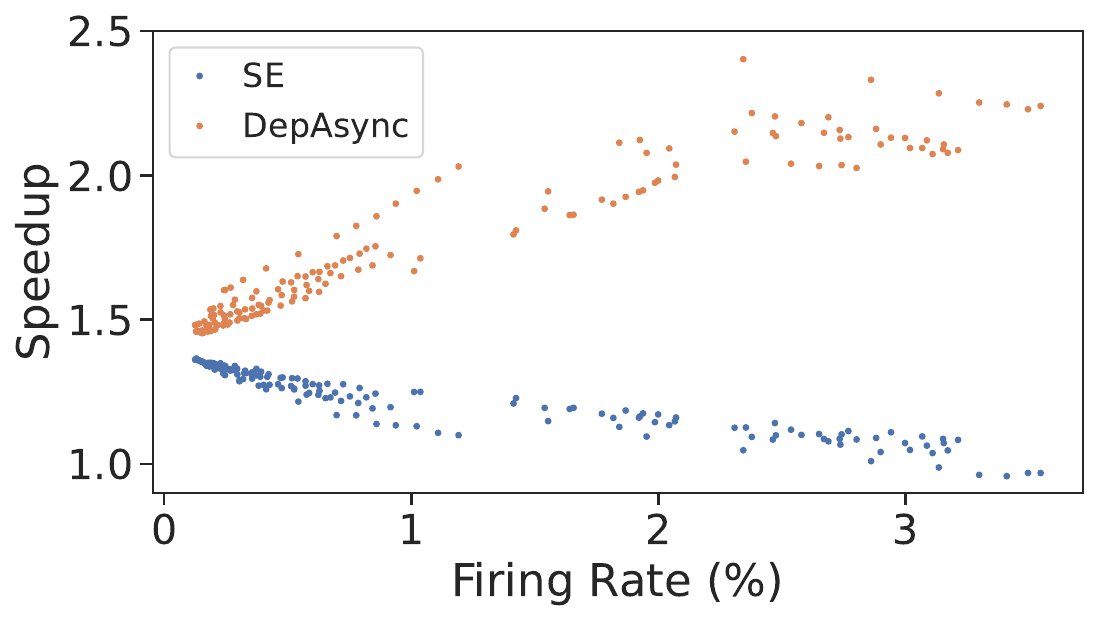}
    }
    \subfloat[Energy efficiency.]{
        \includegraphics[width=0.5\linewidth]{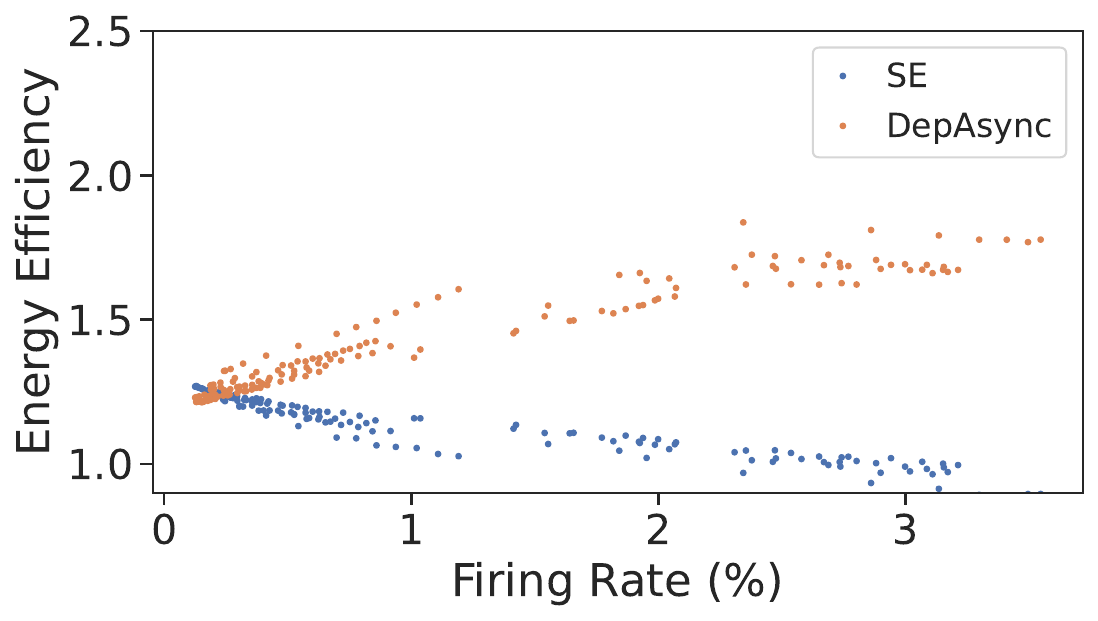}
    }
    \vspace{-0.5em}
    \caption{The impact of the firing rate.}
    \label{fig:fr}
    \vspace{-1.5em}
\end{figure}

Furthermore, we compare SE and DepAsync at different levels of firing rate. As mentioned before, the more spikes are generated, the more likely rollback and recovery (RR) caused by misspeculations occur, which hints at overall performance. We change neuron parameters $\tau_m, V_{rst}$ in Equation~\ref{equ:1} to control the average firing rate in SNN workloads. Fig.~\ref{fig:fr} demonstrates that the more spikes are generated, the more DepAsync exceeds SE, as the performance of SE is limited by the massive RR procedure at a high firing rate. It is worth noting that practical high-performance SNNs either ANN-converted or trained by learning rules behave quite differently from real biological neural networks since the sparse coding and learning mechanism in the human brain is still an open problem. Thus, recent SNNs generally have a higher firing rate, in which case DepAsync provides faster SNN inference performance.

\subsubsection{Cyclic Connections}
\begin{figure}
    \centering
    \subfloat[]{
        \includegraphics[width=0.5\linewidth]{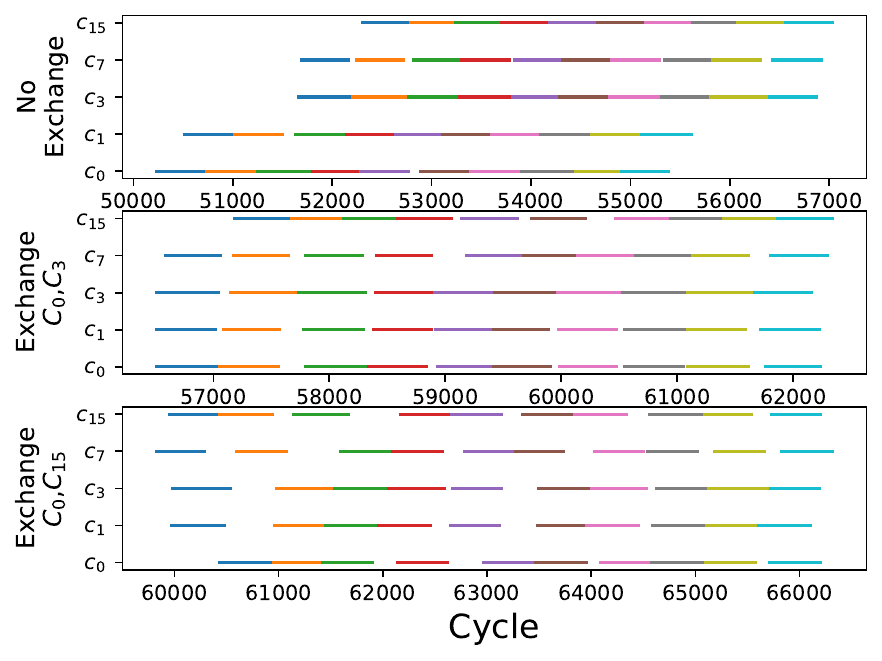}
        \label{fig:run-cyclic}
    }
    \subfloat[]{
        \raisebox{.05\height}{
             \includegraphics[width=0.45\linewidth]{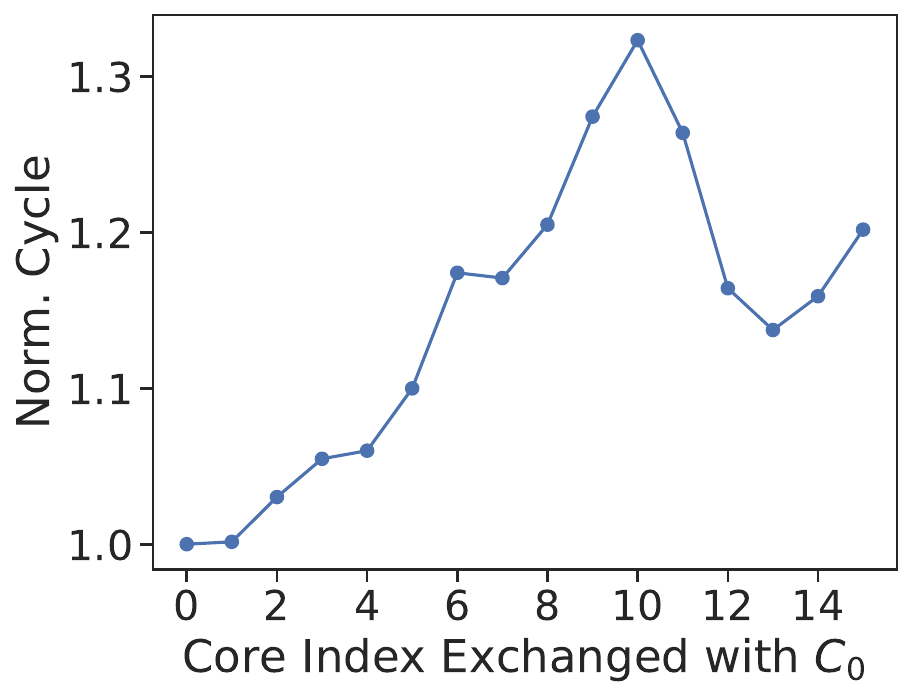}
            \label{fig:speedup-cyclic}
        }
    }
    \vspace{-0.5em}
    \caption{The impact of cyclic connections: (a) A slice and (b) normalized execution time of DepAsync with different cyclic connections.}
    \label{fig:cyclic}
    \vspace{-1em}
\end{figure}

DepAsync analyzes dependencies to control core behavior. When there is a cyclic connection, for example, two cores point at each other, both cores are pre-dependencies to one another. In this case, the post-condition is always satisfied if $m > 1$, however, the pre-condition is broken until all cores finish the current timestep. Thus, the slowest core will trigger others at its end. Finally, the whole system is fallback to a synchronous architecture. We exchange part of neurons between different neuromorphic cores and core $C_0$ to simulate the cyclic connections because $C_0$ holds neurons in the input layer and has no pre-dependencies in the default partition. As shown in Fig.~\ref{fig:run-cyclic}, $C_0$ has to wait for other cores when there are cyclic connections. Fig.~\ref{fig:speedup-cyclic} shows that the performance decreases by 30\% at worst with such cyclic connections. Fortunately, some cyclic connections can be reduced in software compilers with an additional restriction. 

\begin{figure}
    \centering
    \subfloat[Latency.]{
        \hspace*{-1em}
        \includegraphics[width=0.5\linewidth]{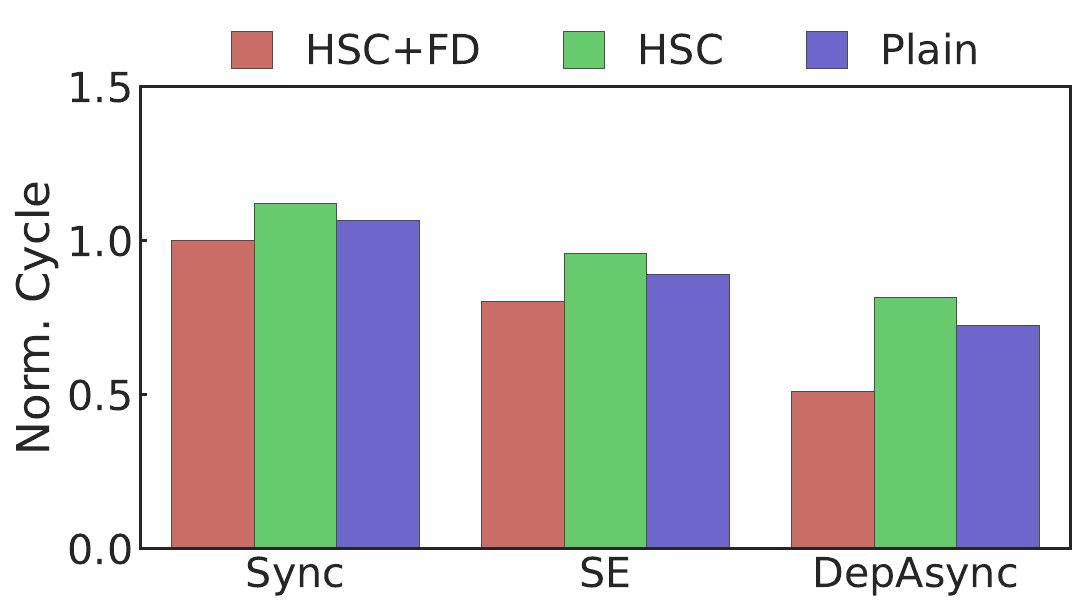}
        \label{fig:speedup-mapping}
    }
    \subfloat[Energy.]{
        \includegraphics[width=0.5\linewidth]{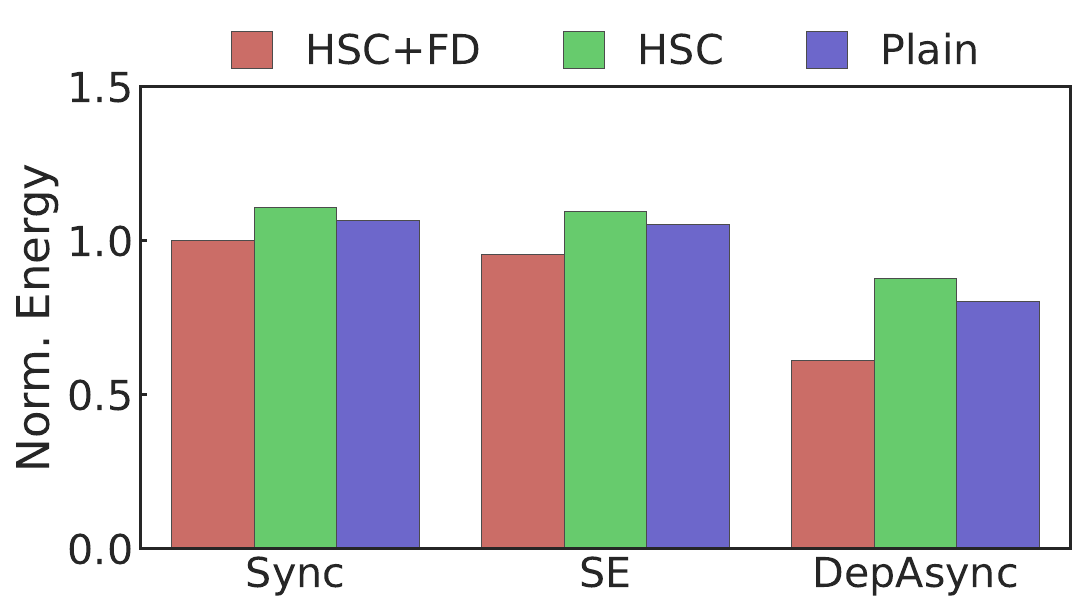}
        \label{fig:energy-mapping}
    }
    \vspace{-0.5em}
    \caption{The impact of mapping algorithms.}
    \label{fig:mapping}
    \vspace{-1.5em}
\end{figure}

\subsubsection{Mapping Algorithms}
The mapping algorithm in SNN compilers is responsible for mapping the logic cores to real physical cores, which affects the distance between a core and its dependencies. Three mapping algorithms are involved in our experiments: Plain means logical cores are orderly mapped to hardware cores; HSC leverages the spatial locality of Hilbert space-filling curves; HSC+FD finetunes HSCs to exploit more locality~\cite{jin2023mapping}. We calculate the average physical distance (measured by NoC hops) between a core
and its dependencies. The results are 2.37, 2.54, and 2.40 hops when using HSC+FD, HSC, and plain mapping algorithms, respectively. The difference in physical distance affects the overall performance. Fig.~\ref{fig:mapping} shows the normalized cycle cost of Sync, SE, and DepAsync with three mapping algorithms. The results demonstrate that DepAsync performs worse than two baselines with poor mapping, which means it is more sensitive to the mapping result. Thus, a better algorithm is necessary for DepAsync. Nevertheless, DepAsync with poor mapping strategies still runs faster and costs less energy than the baseline synchronization architecture.

\subsubsection{NoC Virtual Channels}
\begin{figure}[t]
    \centering
    \subfloat[NoC package blocked cycles.]{
        \hspace*{-1em}
        \includegraphics[width=0.5\linewidth]{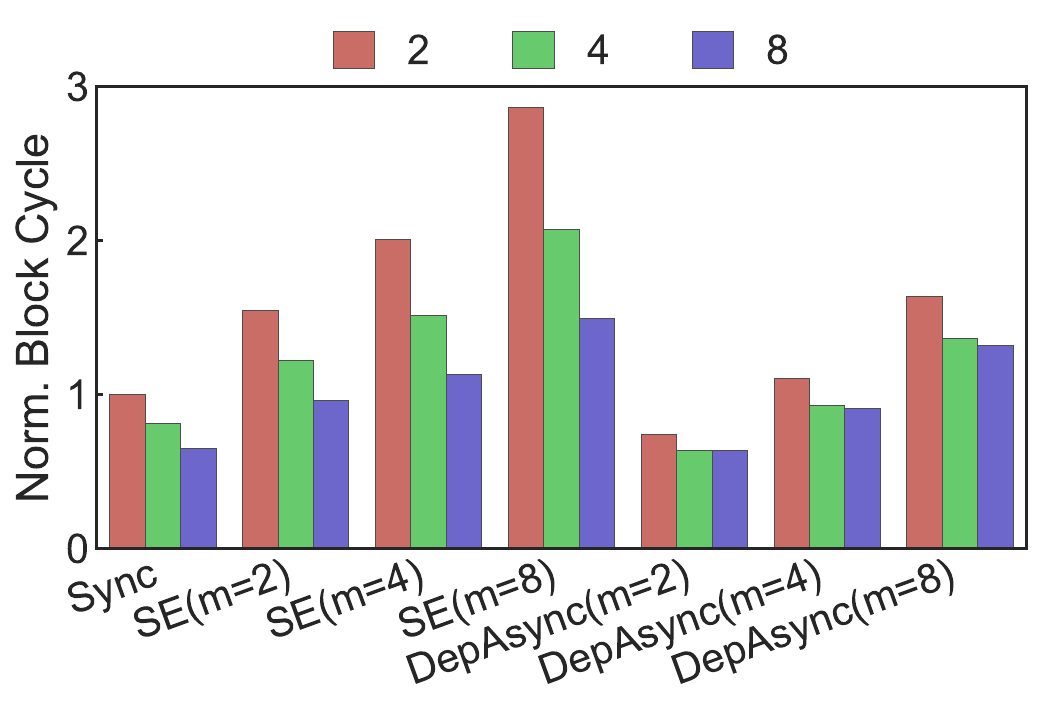}
        \label{fig:block-noc}
    }
    \subfloat[Speedup.]{
        \includegraphics[width=0.5\linewidth]{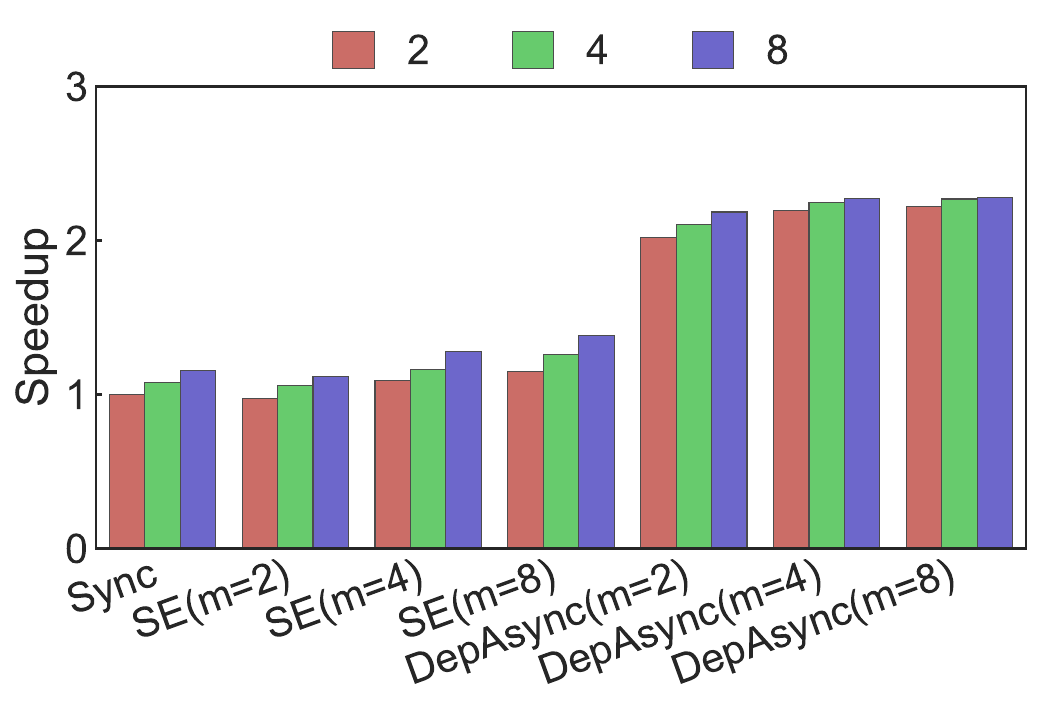}
        \label{fig:speedup-noc}
    }
    \vspace{-0.5em}
    \caption{The impact of NoC virtual channels.}
    \vspace{-1em}
\end{figure}

Fig.~\ref{fig:block-noc} shows the average cycles of spike packages blocked in NoC routers. The block cycle decreases when there are more channels which means the larger NoC bandwidth. The block cycle grows when we increase $m$ in both SE and DepAsync. That is because the more timesteps we allow neuromorphic cores to process, the more spikes will be in-flight at the same time. It is worth noting that the block cycle in DepAsync are smaller than that in Sync when $m=2$. This is because additional packages in DepAsync($m=2$) are from only one future timestep, thus are less than additional packages for global synchronization in Sync. Moreover, block occurs less in DepAsync than in SE with the same $m$. That is because when cores are allowed to process future timesteps, the NoC in DepAsync is only supposed to transfer future spikes, while the NoC in SE has to transfer packages generated by Rollback-and-Recovery operations as well. As for overall performance, Fig.~\ref{fig:speedup-noc} demonstrates speedup due to additional virtual channels, where DepAsync overwhelms Sync and SE in all settings.

\subsection{Scalability of DepAsync}
\begin{figure}[t]
    \centering
    \subfloat[Speedup.]{
        \hspace*{-1.5em}
        \includegraphics[width=0.5\linewidth]{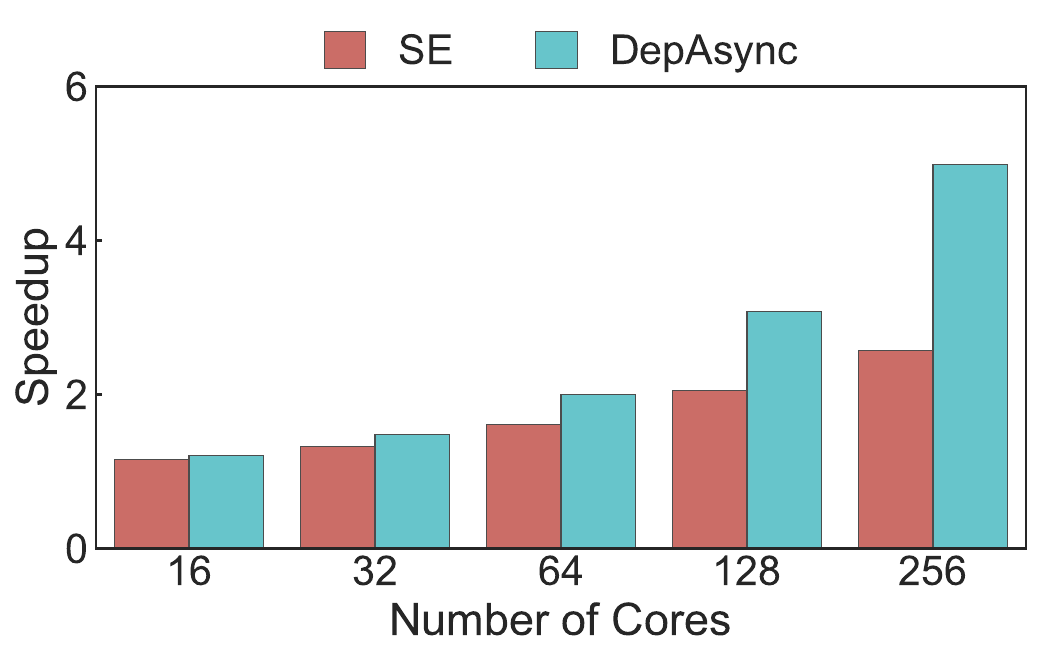}
        \label{fig:scalability-speedup}
    }
    \subfloat[Energy Efficiency.]{
        \includegraphics[width=0.5\linewidth]{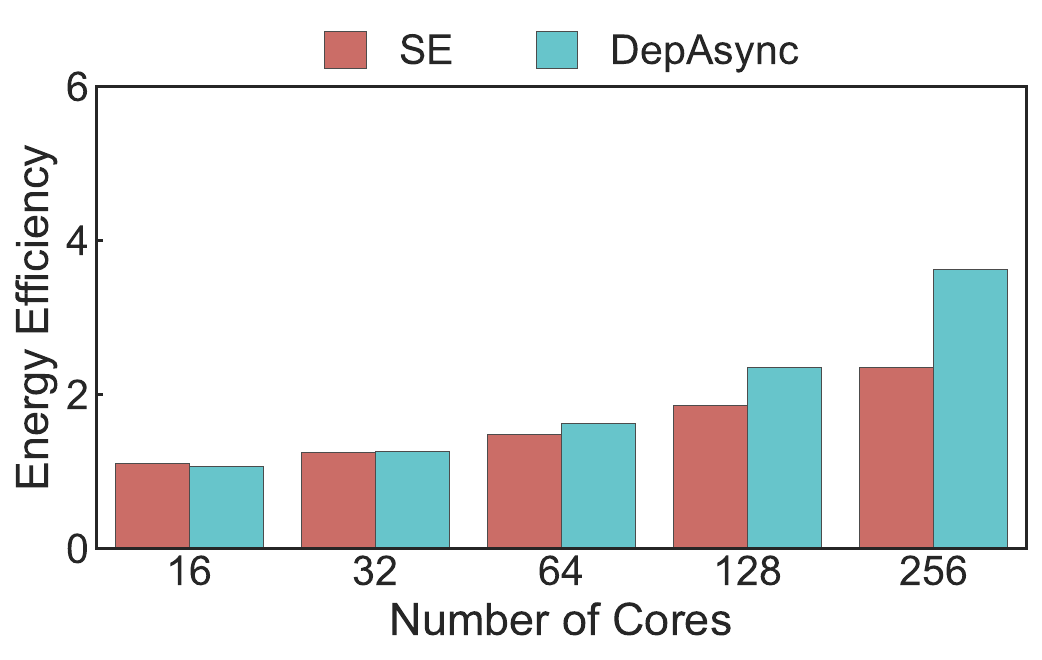}
        \label{fig:scalability-energy}
    }
    \vspace{-0.5em}
    \caption{Scalability.}
    \label{fig:scalability}
    \vspace{-1.2em}
\end{figure}

Finally, we evaluate the scalability of DepAsync. We gradually increase the number of neurons and synapses in the synthetic workload from 16 cores to 256 cores as shown in Table~\ref{tab:scalability}, then evaluate speedup and energy efficiency of DepAsync. The results shown in Fig.~\ref{fig:scalability} identify the excellent scalability of DepAsync compared to Sync. As the system scales up, the synchronization cost becomes more expensive, thus DepAsync accelerates SNN inference faster. The energy efficiency grows as static power consumption decreases due to higher speed. The fact that speedup grows faster than energy efficiency is mainly because spike packages hop in the NoC grows as $O(\sqrt{N_{core}})$ while computation and memory units in neuromorphic cores grow as $O(N_{core})$. DepAsync gains 4.99x speedup and 3.62x energy efficiency at the 256-core scale, which is better than SE.

\begin{table}
    \centering
    \scriptsize
    \caption{Synthetic workloads with different scales}
    \vspace{-0.7em}
    \begin{tabular}{c|c|c}
        \hline
        \textbf{\#Cores} & \textbf{\#Neurons} & \textbf{\#Synapse} \\ \hline \hline
        \makecell{16 ($4\times4$)} & 10,240 & 903,718 \\ \hline
        \makecell{32 ($8\times4$)} & 14,481 & 2,027,922 \\ \hline
        \makecell{64 ($8\times8$)} & 20,480 & 4,048,000 \\ \hline
        \makecell{128 ($16\times8$)} & 28,962 & 8,043,888 \\ \hline
        \makecell{256 ($16\times16$)} & 40,960 & 16,096,000 \\ \hline
    \end{tabular}
    \label{tab:scalability}
    \vspace{-2em}
\end{table}

\begin{figure}[t]
    \centering
    \subfloat[Topology.]{
        \hspace*{-1.5em}
        \includegraphics[width=0.5\linewidth]{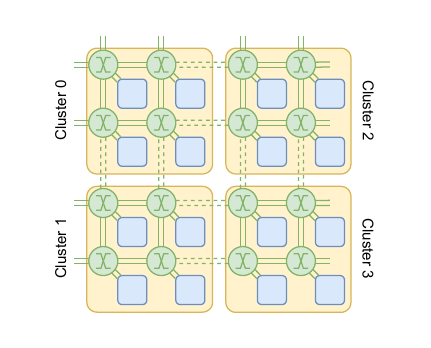}
        \label{fig:bandwidth-topo}
    }
    \subfloat[Latency.]{
        \raisebox{.05\height}{
            \includegraphics[width=0.45\linewidth]{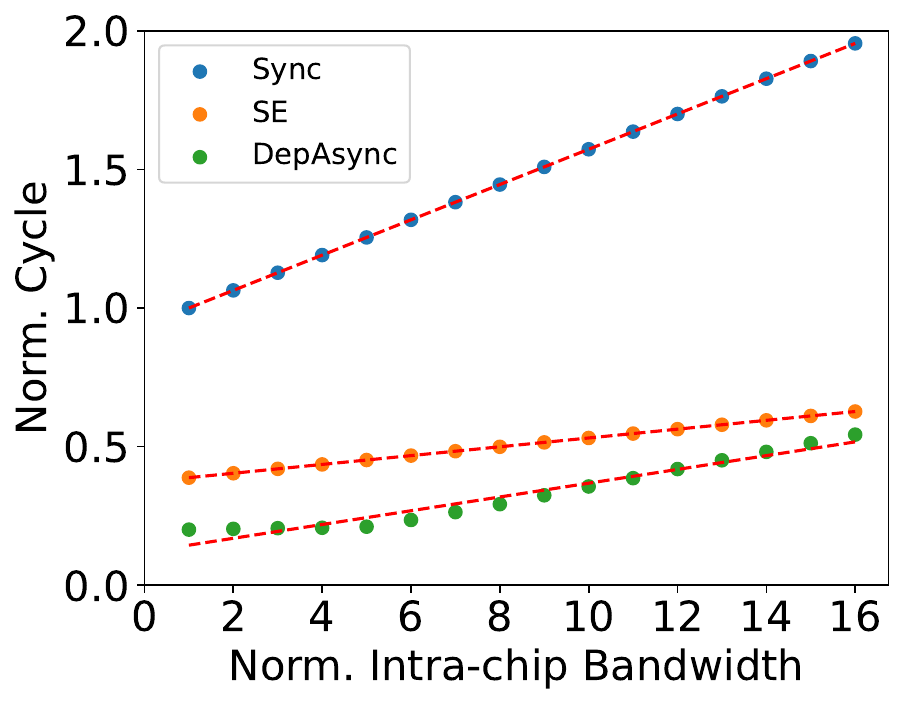}
            \label{fig:bandwidth}
        }
    }
    \vspace{-0.5em}
    \caption{Topology and performance of large scale system with bandwidth hierarchy.}
    \vspace{-1.5em}
\end{figure}

Moreover, we consider the bandwidth hierarchy in real-world many-core systems. Generally, the whole system consists of multiple chips, and each chip contains several neuromorphic cores. The intra-chip NoC is multiple times faster than inter-chip (about 4 times in Darwin~\cite{ma2017darwin}). To simulate the bandwidth hierarchy in a large-scale system, we divide $16\times16$ cores into $8\times8$ 4-core clusters and then decrease the bandwidth between adjacent clusters. Fig.~\ref{fig:bandwidth-topo} gives an example topology of 4 clusters. Fig.~\ref{fig:bandwidth} shows a linear relationship between bandwidth difference and the overall performance.

\subsection{Comparison to ANN-Accelerator-Like Counterparts}
\begin{figure}[t]
    \centering
    \subfloat[Latency.]{
        \hspace*{-1em}
        \includegraphics[width=0.5\linewidth]{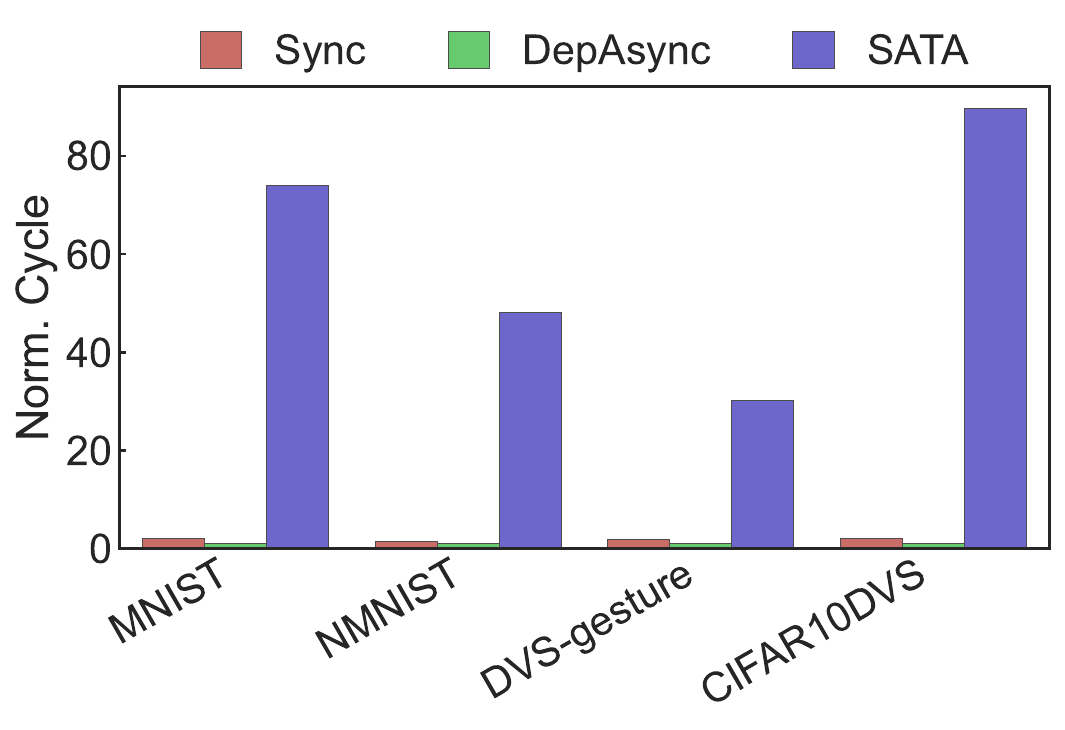}
        \label{fig:speedup-sata}
    }
    \subfloat[Energy.]{
        \includegraphics[width=0.5\linewidth]{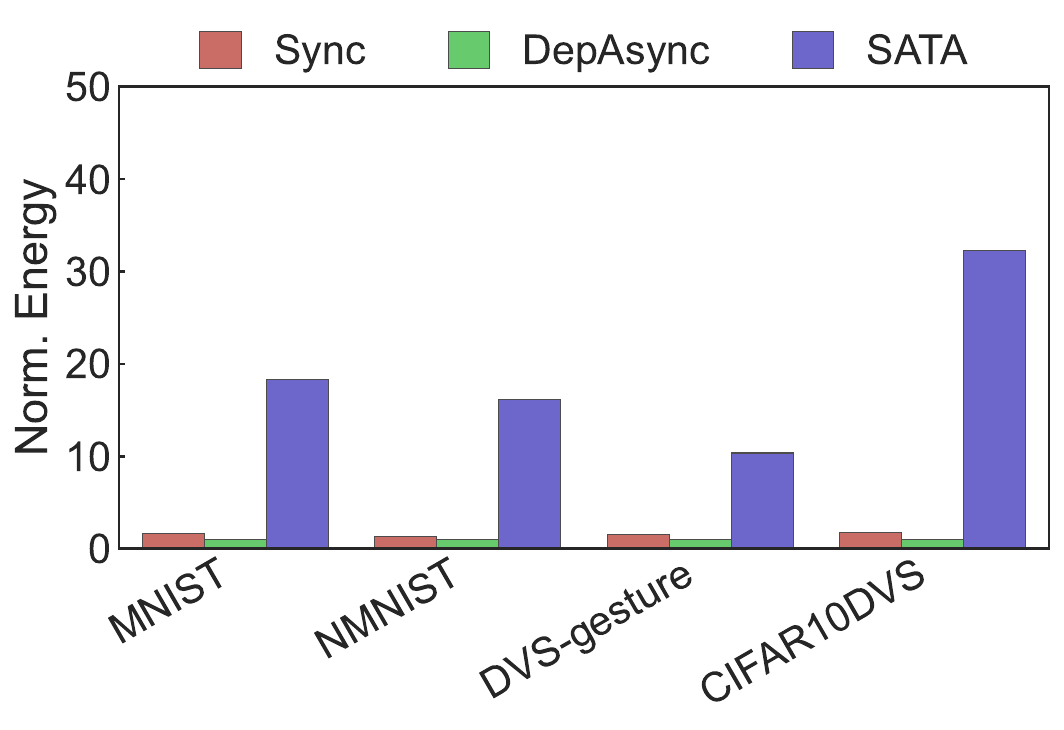}
        \label{fig:energy-sata}
    }
    \vspace{-0.5em}
    \caption{Comparison with the ANN-accelerator architecture.}
    \label{fig:sata}
    \vspace{-1em}
\end{figure}

The ANN-accelerator-like architectures differentiate from our many-core architectures in SNN inference manners. In DepAsync, layers in SNN networks are deployed to different neuromorphic cores and computed simultaneously as mentioned in Section~\ref{sec:many-core}. However, the ANN-accelerator-like architectures act as a layer-sequential mode where a layer will be processed after computing all timesteps of the previous layer, which is not suitable for SNN networks with no clear layer structure ('Synthetic' in Table~\ref{tab:benchmark}). Fig.~\ref{fig:sata} shows latency and energy cost (normalized to DepAsync) in our real-data SNN workloads of the state-of-the-art ANN-accelerator-like architecture SATA~\cite{yin2022sata} which optimized SNN dataflow to reduce data movement energy overhead. Our DepAsync achieves 50.98x speedup and 16.39x energy efficiency on average. The lower energy efficiency is caused by large SRAM overheads in DepAsync.

\section{Conclusions and Discussions}
In this paper, we propose a novel asynchronous architecture to time-accurately accelerate SNN inference on hardware, inspired by dependency relationships on neuromorphic cores. We first observe the imbalance in SNN workloads deployed on multi-core accelerators and identify under-utilization caused by all-core synchronization. Then, we leverage core dependencies determined at compilation time to control the timestep proceeding behavior of neuromorphic cores, implemented by adding a scheduler in each core worked with a new type NoC package. Finally, we use three SNN workloads to evaluate our DepAsync architecture. The experiment results demonstrate that DepAsync brings higher performance and energy efficiency over synchronous accelerators with or without speculative execution, and excellent scalability as scale up.

DepAsync introduces core dependencies as prior information into SNN accelerators. However, other information such as distance between dependencies, synapse numbers of two dependent cores, and dynamic network congestion influences SNN inference performance as well. In future work, more information involved makes more precise schedules which are useful to improve performance. Furthermore, although waiting time of neuromorphic cores is not totally eliminated in DepAsync due to hardware resource limitation, wasted cycles can be utilized by processing another SNN inference if we can deploy multiple SNNs to the same core.


\normalem
\bibliographystyle{IEEEtranS}
\bibliography{refs}

\end{document}